\documentclass[10pt,twocolumn,letterpaper]{article}

\usepackage[pagenumbers]{cvpr} % To force page numbers, e.g. for an arXiv version

\definecolor{cvprblue}{rgb}{0.21,0.49,0.74}
\usepackage[pagebackref,breaklinks,colorlinks,allcolors=cvprblue]{hyperref}
\usepackage{booktabs}
\usepackage{tabularx}
\usepackage{multirow}
\usepackage{makecell}   % multi-line cells with \makecell{...}
\usepackage{tabularx}   % auto-wrapping column type X
\usepackage{caption}    % better caption control (optional)
\usepackage{pifont}              % for \ding
\usepackage[table]{xcolor}       % for colors in tables
\usepackage{pifont}              % for \ding
\usepackage[table]{xcolor}       % for colors in tables
\usepackage{graphicx}
\usepackage{arydshln}
\usepackage{float}

\newcommand{\tick}{\textcolor{green!60!black}{\ding{51}}} % ✓
\newcommand{\cross}{\textcolor{red!70!black}{\ding{55}}}   % ✗

\title{Know-Show: Benchmarking Video-Language Models on Spatio-Temporal Grounded Reasoning}

\author{Chinthani~Sugandhika$^{1,2,3}$~\and~Chen~Li$^{2,3}$~\and~Deepu~Rajan$^{1}$~\and~Basura~Fernando$^{1,2,3}$ \\
$^{1}$College of Computing and Data Science, Nanyang Technological University, Singapore\\
$^{2}$Institute of High-Performance Computing, Agency for Science, Technology and Research, Singapore\\
$^{3}$Centre for Frontier AI Research, Agency for Science, Technology and Research, Singapore
}

\begin{document}
\maketitle
\begin{abstract}

Large Video-Language Models (Video-LMs) have achieved impressive progress in multimodal understanding, yet their reasoning remains weakly grounded in space and time. We present Know-Show, a new benchmark designed to evaluate spatio-temporal grounded reasoning, the ability of a model to reason about actions and their semantics while simultaneously grounding its inferences in visual and temporal evidence. Know-Show unifies reasoning and localization within a single evaluation framework consisting of five complementary scenarios across spatial (person, object, person–object, and hand–object) and temporal dimensions. Built from Charades, Action Genome, and Ego4D with 2.5K high-quality human-authored questions, the benchmark exposes significant gaps between current Video-LMs and human reasoning. To bridge this gap, we propose GRAM, a training-free plug-in that augments Video-LMs with fine-grained grounded reasoning through attention-based video token selection and explicit timestamp encoding. Extensive experiments across open and closed Video-LMs (Qwen, VideoR1, Gemini and GPT-4o etc.) reveal that existing models struggle to “show what they know" and vice versa. Know-Show establishes a unified standard for assessing grounded reasoning in video-language understanding and provides insights toward developing interpretable and reliable multimodal reasoning systems. 
We have released the dataset at \href{https://github.com/LUNAProject22/Know-Show}{https://github.com/LUNAProject22/Know-Show}, and the code will be released in the same repository.

% \noindent \textbf{Keywords}: Video-Language models. Grounded reasoning. Spatio-temporal understanding

% , while GRAM consistently improves spatio-temporal grounding without retraining. 

% \todo{Basura: This needs to be shorted and to the point. No need to elaborate too much. can also go to abstract..}
% We evaluated our benchmark on xx video-LMs and observed that they struggle with this unified task.
% This difficulty arises because reasoning and grounding have traditionally been treated as independent problems, with training objectives and evaluation benchmarks evolving along separate trajectories. As a result, existing models demonstrate only partial success on our dataset as shown in Fig. \ref{fig:intro_diagram}. In essence, these models tend to “know” without being able to “show”.
% Thus, only models that have spatio-temporal grounded reasoning capability can perform well on the proposed benchmark.  
% In particular, they should be able to attend to the correct action while reasoning and to generate interpretable evidence such as bounding boxes or timestamps aligned with their predictions.

\end{abstract}    
\section{Introduction}
\label{sec:intro}

\begin{figure}[htbp]
  \centering
  \includegraphics[width=1\columnwidth]{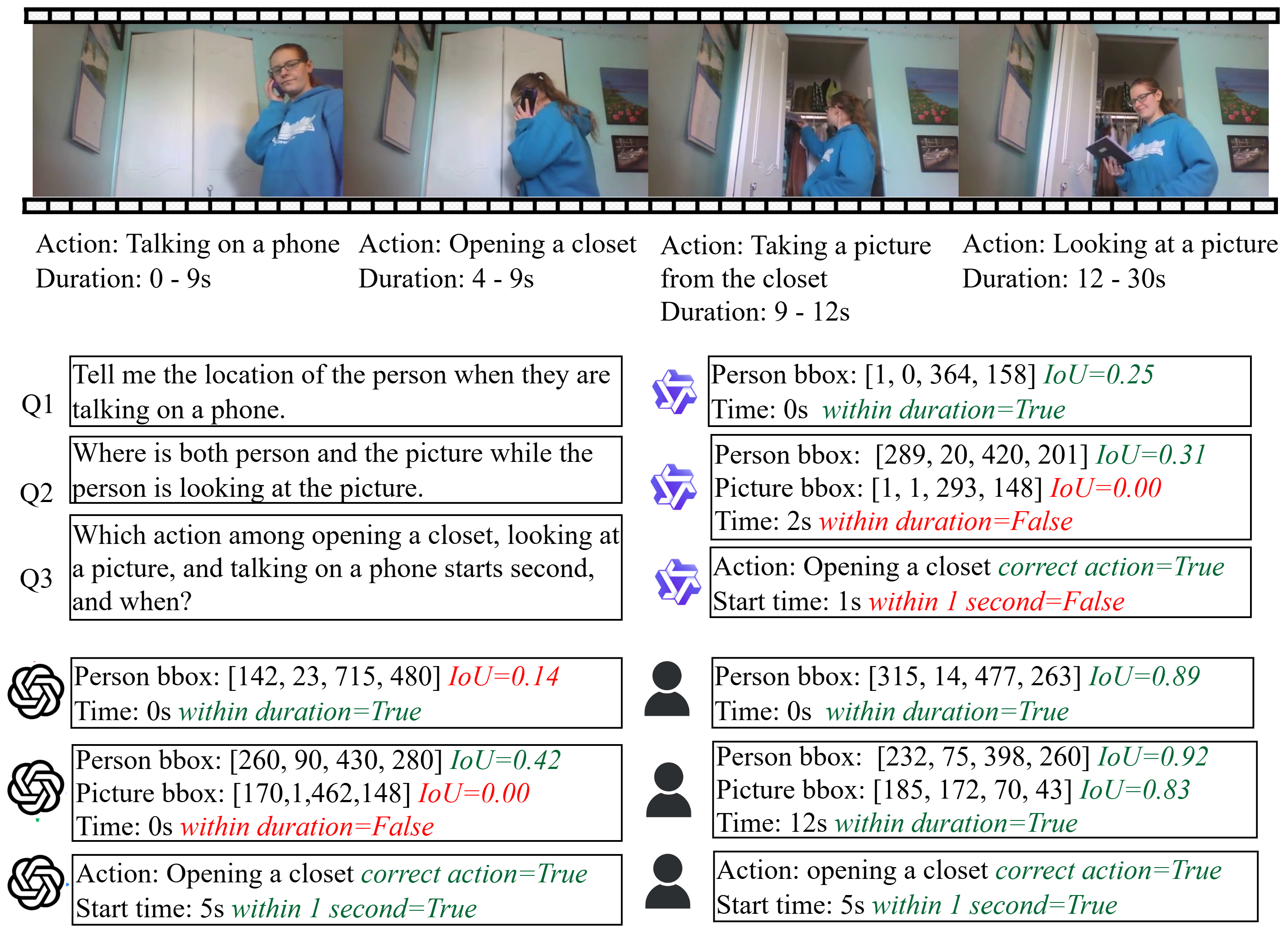}
   \caption{Challenges of current Video-Language Models (\includegraphics[height=0.8em]{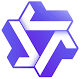} Qwen \includegraphics[height=0.8em]{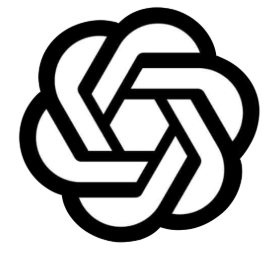} GPT-4o \includegraphics[height=0.8em]{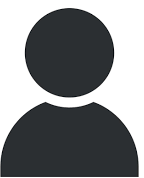} Human) in \emph{spatio-temporal grounded reasoning}.
    Although models can capture coarse temporal order of actions or recognize objects, they often fail to correctly associate actions, humans, and objects with their precise spatial and temporal contexts.    
    Unlike humans, who can reason as well as ground, most models cannot do both. 
    We propose the \textbf{Know-Show} benchmark, to assess genuine spatio-temporal grounded reasoning, a crucial capability for real-world applications such as robotics and embodied AI.}
  
  \label{fig:intro_diagram}
\end{figure}

% Recent advances in Large Language Models (LLMs) have revolutionized the landscape of artificial intelligence, exhibiting impressive capabilities in language understanding, reasoning, and planning across diverse NLP tasks.
% Building on this, Large Multimodal Models such as CLIP \cite{clip}, LLaVA \cite{li2023llava}, and BLIP-2 \cite{li2023blip} extend language models to the visual domain by bridging images and text.
% The next frontier naturally extends from static images to dynamic video, where spatio-temporal reasoning becomes central. This evolution has led to the emergence of Video-Language Models (Video-LMs) such as LLaVA-OneVision \cite{llavaonevision}, Video-LLaVA \cite{videollava}, Qwen \cite{qwen}, and Gemini \cite{gemini}, which can interpret, reason about, and converse over videos in natural language enabling broad applications such as robotics, autonomous driving, embodied and assistive AI etc.

Recent progress in large multimodal models has extended the reasoning capabilities of Large Language Models (LLMs) from text to vision. Building on systems such as CLIP \cite{clip}, LLaVA \cite{li2023llava}, and BLIP-2 \cite{li2023blip}, Video-Language Models (Video-LMs) including Qwen \cite{qwen}, VideoChat2 \cite{li2024mvbench}, VideoR1 \cite{feng2025video}, Gemini \cite{gemini}, and GPT-4o \cite{gpto} enable models to interpret and converse about videos in natural language, supporting broad array of applications in robotics, autonomous driving, security and surveillance, embodied AI, and assistive technologies.

% Ability to see is not the same as the ability to understand and reason. 
% Grounded visual reasoning requires identifying entities, their spatial relations, and temporal dependencies, and using these structured relationships to support inference. 
Grounded visual reasoning requires more than recognizing objects or actions, it demands identifying entities, modeling their relationships, tracking their temporal evolution, and anchoring inferences directly to where and when they occur.
Cognitive science suggests that humans naturally abstract visual input into relational scene structures (e.g., “A is left of B,” “X occurs before Y”) with appropriate spatial and temporal grounding, which form the basis for grounded visual reasoning \cite{lake2017building,johnson2017inferring}. These abstractions enable humans to reason in a grounded manner, linking what is inferred to where and when it occurs.
As Video-LMs are increasingly deployed in real-world settings, it is crucial to assess whether they possess this grounded reasoning capability. 
Yet existing benchmarks fall into two largely disjoint categories as localization and reasoning centric. Localization-centric datasets (e.g., Action Genome \cite{ag}, VidVRD \cite{shang2017video}, Ego4D \cite{ego4d}, Kinetics \cite{kay2017kinetics}, ActivityNet \cite{caba2015activitynet}) emphasize spatial and temporal annotation but focus on recognition rather than reasoning. Reasoning-centric benchmarks (e.g., MV-Bench \cite{li2024mvbench}, Vinoground \cite{zhang2024vinoground}, TempCompass \cite{liu2024tempcompass}, STAR \cite{star}), summarized in Table~\ref{tab:benchmark_comparison}, evaluate compositional reasoning but do not essentially require grounding predictions in the video.
% \red {Some closely related work such as  evaluate either} 
% As a result, current Video-LMs may produce fluent explanations that appear plausible but are not supported by visual evidence, instead relying on language priors and common-sense correlations \cite{nextgqa}.

Motivated by this gap, we pose a simple yet fundamental question: “Are today’s video-language models good at performing grounded reasoning? In other words, do they show what they know and do they know what they show?
% aka \emph{showing what they know and knowing what they show}”. 
To answer this, we introduce a new benchmark designed to evaluate the spatio-temporal grounded (video) reasoning capability of Video-LMs by unifying reasoning and localization within a single, cohesive framework. 
Spatio-temporal grounded reasoning is the capability of a model to perform high-level reasoning about actions and their semantics (i.e., humans, objects, hands and interactions) in videos while simultaneously grounding their inferences in the corresponding spatial and temporal dimensions. Unlike reasoning tasks that emphasize abstract inference without verification, or grounding tasks that focus solely on perceptual localization, spatio-temporal grounded reasoning requires models to both know and show i.e., to infer the correct outcome and to justify it by providing the relevant spatial and temporal cues in the video.

% \red{ Action–role binding between interacting entities refers to the ability of a model to associate an action with all participating entities by assigning each entity its correct semantic role, while grounding these bindings consistently across spatial and temporal dimensions.
% A key mechanism underlying spatio-temporal grounded reasoning is action–role binding between interacting entities, which refers to the ability of a model to associate an action with all participating entities by assigning each entity its correct semantic role and grounding these bindings consistently across spatial and temporal dimensions.
% Although xxxx involve spatio-temporal grounding and event reasoning, they do not require action–role binding, as the actions involve only a single entity and do not necessitate assigning complementary semantic roles among interacting entities.
% Action-conditioned spatio-temporal grounding requires models to infer the relevant action prior to localizing entities, while action–role binding further requires assigning correct semantic roles to multiple interacting entities during spatial grounding.
% }
% This evaluates whether a model can jointly reason and localize entities by binding each participant in an interaction (e.g., objects, hands, or people) to its correct semantic role in the action, while maintaining coherent spatial and temporal grounding of these role assignments.

Accordingly, as illustrated in Fig.~\ref{fig:test_Cases}, our benchmark consists of five complementary test scenarios grouped into two categories:
(a) \textbf{Action-Conditioned Spatial Grounded Reasoning}, 
which evaluates a model’s ability to jointly reason and localize objects, hands, people, or their interactions within specific action contexts; and 
% It requires action–role binding, i.e., associating the action with its participating entities; and
(b) \textbf{Action-Conditioned Temporal Grounded Reasoning}, which assesses whether the model can
reason about actions over time, including temporally localizing actions and determining their relative ordering.
A model receives credit only when it succeeds in both reasoning and localization, demonstrating a true “know-and-show” capability. 
Through this unified formulation, our benchmark bridges the traditional divide between reasoning and localization, offering a principled evaluation of spatio-temporal grounded reasoning, a core requirement for reliable and interpretable video understanding.
These reasoning scenarios collectively probe a set of fundamental reasoning and localization capabilities, including accurate action recognition, precise action localization, relative temporal order reasoning, action–role binding (associating the action with its participating entities), action-conditioned spatial grounding, compositional spatio-temporal reasoning, and fine-grained interaction reasoning. Together, these competencies form the foundation of robust spatio-temporal grounded reasoning in videos.
Such spatio-temporal grounded reasoning is essential for real-world applications where reasoning, interpretability, and grounding are closely interconnected.
For example, in smart homes and elder care, verifying safety-critical events requires both spatial and temporal evidence, such as confirming whether a person actually turned off the stove by showing the position of the knob and the corresponding time.
In the same way, robotics, embodied AI, surveillance, and safety or compliance monitoring rely on queries that integrate logical reasoning with grounded spatial and temporal evidence.

We evaluated our benchmark on multiple video-LMs and found that they struggle with this unified task. As shown in Fig. \ref{fig:intro_diagram}, their reasoning remains weakly grounded in visual evidence and often misaligned in time \cite{nextgqa}.
% To address this, we introduce GRAM (Grounded \textbf{R}easoning \textbf{A}ugmentation \textbf{M}odule), a training-free, plug-and-play baseline framework that enhances any Video-LM with fine-grained spatio-temporal grounded reasoning capabilities. GRAM grounds each reasoning step in the most relevant video evidence by aggregating attention across layers to select the most similar video tokens and fusing them with the input tokens during decoding. Additionally, we incorporate an explicit timestamp interleaving scheme to improve temporal alignment. Together, these components enable interpretable, step-by-step reasoning grounded directly in the observed video content.
% 
To address this, we introduce GRAM (\textbf{G}rounded \textbf{R}easoning \textbf{A}ugmentation \textbf{M}odule), a training-free, plug-and-play baseline framework that equips Video-LMs with fine-grained spatio-temporal grounded reasoning. GRAM grounds each reasoning step in the most relevant video evidence by aggregating attention across layers to retrieve the most relevant video tokens and fusing them into the decoding process. We further incorporate an explicit timestamp-interleaving scheme to improve absolute temporal alignment. Together, these components enable interpretable reasoning directly grounded in the video.

% Building on these observations, we propose a training-free, plug-and-play baseline framework called \emph{GRAM} that augments any Video-LM with fine-grained spatio-temporal grounded reasoning capabilities. 
% Unlike typical video-LMs that encode the video, fuse these with textual embeddings, and perform reasoning based on this global representation of text and video,
% GRAM explicitly grounds each reasoning step in the most relevant video evidence.
% It aggregates attention across all layers of the underlying VLM, selects the most similar video tokens to the current input, and fuses them with input tokens at each decoding step.
% This ensures that predictions are guided by spatially and temporally localized cues.
% Furthermore, to improve temporal alignment, we introduce an explicit timestamp interleaving scheme, providing consistent absolute time references across videos. 
% Together, these components enable step-by-step reasoning that is both interpretable and grounded in the observed video content.
% Furthermore, to improve temporal alignment, we introduce a dense ordinal temporal encoding followed by an explicit timestamp interleaving scheme, providing consistent and semantically meaningful absolute time references across videos. Together, these components enable step-by-step reasoning that is both interpretable and grounded in the observed video content.

In summary, our contributions are three-fold.
First, we introduce the Know-Show Benchmark, 
that unifies reasoning and grounding to assess the spatio-temporal grounded reasoning capability in Video-LMs.
Second, we conduct a comprehensive evaluation of a series of open and closed Video-LMs on this benchmark, revealing systemic weaknesses in their ability to align reasoning with visual evidence across space and time.
Third, we present GRAM, a simple, training-free augmentation that enhances this capability of video-LMs.
Our experiments reveal following important findings: 
(a) humans naturally exhibit strong spatio-temporal grounded reasoning, achieving over 75\% accuracy on our benchmark.
(b) despite recent advances, current Video-LMs remain far from human-level on this task,
(c) while these models handle temporal and person grounded reasoning tasks reasonably well, they show substantial limitations in object and person–object co-grounded reasoning scenarios, and
(d) finally, all models almost fail on hand–object co-grounded reasoning, where precise, fine-grained hand-object contact interactions are essential.

\section{Previous Work}
\label{sec:previous_work}

\begin{table*}[tb]

\centering
\scriptsize
\resizebox{\textwidth}{!}{
\begin{tabular}{|l| c| c| c| c| c| c| c| c| c| c| c|}
\hline
Dataset & Type & \#Q & Comp. & TGR & PGR & OGR & POCoGR & HOCoGR & Test Src. & Human Eval. & Domain \\
\hline

AGQA \cite{grunde2021agqa} \textcolor{gray}{CVPR'21} & OQA,MCQ & 39M & \tick & \cross & \cross & \cross & \cross & \cross & T,SG & \tick & Open (ActionGenome +1) \\

STAR \cite{star} \textcolor{gray}{NIPS'21} & MCQ & 7K & \tick & \cross & \cross & \cross & \cross & \cross & H,T,SG & \cross & Opne (ActionGenome +1) \\

ContrastSets \cite{park2022exposing} \textcolor{gray}{NAACL'22} & MCQ & 1K & \tick & \cross & \cross & \cross & \cross & \cross & H,T,LLM & \tick & Mixed (MSR-VTT +1) \\

Vid-STG \cite{zhang2020does} \textcolor{gray}{CVPR'20} & OQA & 9K & \tick & \cross & \tick & \tick & \cross & \cross & SG & \cross & Open( VidOR) \\

% HC-STVG \cite{tang2021human} \textcolor{gray}{TCSVT'21} & STVG & 5.6K & \tick & \tick & \tick & \cross & \cross & \cross & H & \cross & Open (AVA) \\

TestOfTime \cite{bagad2023test} \textcolor{gray}{CVPR'23} & BQA & 1.8K & \tick & \tick & \cross & \cross & \cross & \cross & T & \cross & Open (TEMPO +2) \\

PerceptionTest \cite{patraucean2023perception} \textcolor{gray}{NIPS'23} & MCQ & 6K & \tick & \cross & \cross & \cross & \cross & \cross & H & \tick & Indoor (Manual) \\

% VideoCon \cite{bansal2024videocon} \textcolor{gray}{CVPR'24} & MCQ & 305K & \cross & \cross & \cross & \cross & \cross & \cross & H,LLM & \tick & Movies (MovieClips) \\

Cinepile \cite{rawal2024cinepile} \textcolor{gray}{CVPR'24} & BQA,OQA & 38K & \tick & \cross & \cross & \cross & \cross & \cross & H,LLM & \cross & Open (MSR-VTT +2) \\

SEED-Bench \cite{li2024seed} \textcolor{gray}{CVPR'24} & MCQ & 24K & \cross & \cross & \cross & \cross & \cross & \cross & H,LLM & \cross & Open (Charades +2) \\

Next-GQA \cite{nextgqa} \textcolor{gray}{CVPR'24} & MCQ & 10.5K & \tick & \cross & \cross & \cross & \cross & \cross & H & \tick & Open (NExT-QA) \\

MV-Bench \cite{li2024mvbench} \textcolor{gray}{CVPR'24} & MCQ & 4K & \tick & \cross & \cross & \cross & \cross & \cross & T,LLM & \cross & Open (MoVQA +10) \\

TempCompass \cite{liu2024tempcompass} \textcolor{gray}{ACLFin.’24} & MCQ,BQA & 7.5K & \tick & \cross & \cross & \cross & \cross & \cross & H,LLM & \tick & Open (ShutterStock) \\

MMBench-Video \cite{fang2024mmbench} \textcolor{gray}{NIPS'24} & OQA & 2K & \tick & \cross & \cross & \cross & \cross & \cross & H & \cross & Open (YouTube) \\

VITATECS \cite{li2024vitatecs} \textcolor{gray}{ECCV'24} & BQA & 13K & \tick & \cross & \cross & \cross & \cross & \cross & H,LLM & \tick & Open (MSRVTT +1) \\

Next-GQA \cite{nextgqa} \textcolor{gray}{CVPR'24} & MCQ & 10.5K & \tick & \tick & \cross & \cross & \cross & \cross & H & \tick & Open (NExT-QA) \\

Vinoground \cite{zhang2024vinoground} \textcolor{gray}{arXiv-24} & BQA & 1K & \tick & \cross & \cross & \cross & \cross & \cross & H,LLM & \tick & Open (VATEX) \\

VideoVista \cite{li2024videovista} \textcolor{gray}{arXiv-24} & MCQ & 25K & \cross & \cross & \cross & \cross & \cross & \cross & T,LLM & \cross & Mixed (Panda-70M) \\

VELOCITI \cite{saravanan2025velociti} \textcolor{gray}{CVPR'25} & BQA & 3.1K & \tick & \cross & \cross & \cross & \cross & \cross & H,T,LLM & \tick & Movies (VidSitu) \\

SVLTA \cite{du2025svlta} \textcolor{gray}{CVPR'25} & OQA & 25.3K & \tick & \cross & \cross & \cross & \cross & \cross & T & \cross & Synthetic videos \\

OVO-Bench \cite{niu2025ovo} \textcolor{gray}{CVPR'25} & OQA & 2.8K & \tick & \cross & \cross & \cross & \cross & \cross & H, LLM & \tick & Open (OpenEQA +10) \\

VIDHALLUC \cite{VIDHALLUC} \textcolor{gray}{CVPR'25} & BQA & 8.2K & \tick & \cross & \cross & \cross & \cross & \cross & T, H & \tick & Open (ActivityNet +2) \\

VideoHallu \cite{li2025videohallu} \textcolor{gray}{NIPS'25} & OQA & 3K & \tick & \cross & \cross & \cross & \cross & \cross & H, LLM & \cross & Synthetic videos \\

MVU-Eval \cite{MVU-Eval} \textcolor{gray}{NIPS'25} & OQA & 4.9K & \tick & \cross & \cross & \cross & \cross & \cross & T, H & \cross & Open (Kinetics-400 +7) \\

% ST-Align \cite{li2025llava} \textcolor{gray}{CVPR'25} & STVG & 2K & \tick & \tick & \tick & \tick & \cross & \cross & LLM & \cross & Open \\

CVRR-ES \cite{khattak2025good} \textcolor{gray}{arXiv-25} & OQA & 2.4K & \cross & \tick & \cross & \cross & \cross & \cross & H,LLM & \tick & Open (SSV2, CATER, +5) \\

Video-MME \cite{fu2025video} \textcolor{gray}{arXiv-25} & MCQ & 2.7K & \cross & \tick & \cross & \cross & \cross & \cross & H & \cross & Open(YouTube) \\

VideoMolmo \cite{ahmad2025videomolmo} \textcolor{gray}{arXiv'25} & OQA & 0.1K & \tick & \cross & \tick & \tick & \cross & \cross & H & \cross & Open (nuScenes, +4) \\

% OmniGround \cite{gao2025omniground} \textcolor{gray}{arXiv'25} & STVG & 3.5K & \tick & \tick & \tick & \tick & \cross & \cross & H & \cross & Open (internet videos) \\

R-AVST \cite{li2025llava} \textcolor{gray}{arXiv'25} & OQA & 8K & \tick & \cross & \tick & \tick & \cross & \cross & LLM & \cross & Open (UnAV-100) \\

\hline
\rowcolor{gray!15} Ours & OQA & 2K & \tick & \tick & \tick & \tick & \tick & \tick & H & \tick & Open (Ego4D +2) \\

\hline
\end{tabular}
}
\caption{
Comparison of existing video-language reasoning benchmarks with our proposed \textbf{Know-Show Benchmark}.
Our benchmark provides a unified evaluation of reasoning and grounding through \emph{Action-Conditioned Spatio-Temporal Grounded Reasoning}.
Abbreviations:
Task - MCQ: Multiple Choice, OQA: Open-ended Question-Answering, BQA: Binary Question-Answering; 
% STVG: Spatio-Temporal Video Grounding;
\#Q - Number of questions;
Comp. - Compositionality;
TGR - Action-conditioned temporal grounded reasoning;
PGR - Action-conditioned person grounded reasoning;
OGR - Action-conditioned object grounded reasoning;
POCoGR - Action-conditioned person-object co-grounded reasoning;
HOCoGR - Action-conditioned hand-object co-grounded reasoning;
Test Src. - Source used for test question creation (H: Human, T: Template, LLM: Large Language Model, SG: Scene Graph);
Human Eval. - Indicates whether human evaluation was performed;
Domain - Open: natural videos, Movies, Mixed: natural \& movies, Synthetic: artificially generated videos.
}
\label{tab:benchmark_comparison}
\end{table*}

\noindent \textbf{Video-Language Benchmarks.}
% MOMA-QA- box augmented questions
The benchmarks broadly related to our work are presented in Table~\ref{tab:benchmark_comparison}. 
Amongst, datasets such as VITATECS \cite{li2024vitatecs}, MMBench \cite{fang2024mmbench}, TempCompass \cite{liu2024tempcompass}, and STAR \cite{star} includes high-level reasoning over temporal and/or spatial relationships, they do not require specifically to ground their inferences on space or time,
% . As a result, models can often produce plausible answers without demonstrating that their reasoning is supported by visual evidence.
while datasets like MV-Bench \cite{li2024mvbench}, Video-MME \cite{fu2025video}, VideoVista \cite{li2024videovista} include question categories that assess the model’s ability to directly locate actions and objects.
% in space or time.
Consequently, there remains a lack of benchmarks designed to evaluate whether models can reason about actions and their semantics while simultaneously identifying the corresponding spatial and temporal evidence in the video. To address this gap, we introduce a unified evaluation framework that requires models to jointly reason and ground across space and time.
Our benchmark encompasses four spatial grounded reasoning scenarios: action-conditioned person, object, person–object, and hand–object reasoning, in increasing order of compositional complexity, along with action-conditioned temporal grounded reasoning.
A closely related effort is NExT-GQA \cite{nextgqa}, which targets temporally grounded reasoning but uses MCQ format that is more susceptible to language priors and vision–language shortcuts. 
% In contrast, our benchmark evaluates both spatial and temporal grounded reasoning and adopts open-ended QA, reducing the biases inherent in MCQs. 
% 
% Another related line of work \cite{zhang2020does, ang2021human, li2025llava, ahmad2025videomolmo} addresses Spatio-Temporal Video Grounding, which aims to detect and track the spatio-temporal tube of a queried object given a natural language phrase. These approaches primarily focus on localization and tracking, and do not necessarily require reasoning about relative action order or action semantics.
% 
Vid-STG \cite{zhang2020does}, VideoMolmo \cite{ahmad2025videomolmo} and R-AVST \cite{li2025llava} includes some spatially grounded queries but are limited to either a person or an object, but they do not evaluate co-grounded reasoning across multiple interacting entities (e.g., person–object or hand–object). As a result, these benchmarks do not require explicit action–role binding among interacting entities, nor do they assess whether models can reason over multi-entity interactions. Importantly, none of the prior benchmarks include dedicated evaluation scenarios for fine-grained hand–object co-grounded reasoning. Such fine-grained interactions are critical in real-world applications, including robotics and embodied manipulation where understanding precise hand–object dynamics is essential for reliable decision-making.

\noindent \textbf{Spatio-Temporal Grounded Reasoning for Video-LMs.}
Most state-of-the-art Video-LMs are transformer-based \cite{vaswani2017attention} models pretrained on large-scale image–text \cite{lei2021less}, video–text \cite{li2020hero}, or hybrid \cite{fu2023empirical} data, often built upon strong language backbones such as LLaMA \cite{touvron2023llama}. Although they achieve strong performance on video question answering, they primarily optimize answer accuracy rather than reasoning faithfulness \cite{xiao2024can}, leading to attention patterns that reflect incidental correlations rather than grounded visual evidence \cite{nextgqa}.
Inspired by chain-of-thought prompting, recent multimodal CoT approaches (e.g., CCoT \cite{mitra2024compositional}, DDCoT \cite{zheng2023ddcot}, SCAFFOLD \cite{lei2024scaffolding}, ICOT \cite{gao2025interleaved}) introduce intermediate reasoning steps via scene graphs or object annotations, yet they largely produce text-only explanations. More recent evidence-centric methods \cite{meng2025open, fan2025grit, wang2025traceable, zheng2025deepeyes} explicitly operate on visual regions (e.g., cropping or zooming) to gather supporting evidence,
% and LRR \cite{bhattacharyya2023look} augments video reasoning via surrogate grounding tasks in the training pipeline.
While these works reflect a growing shift toward grounded reasoning, most focus on images e.g., ICoT \cite{gao2025interleaved} or require training-intensive pipelines e.g., LRR \cite{bhattacharyya2023look}. In contrast, our framework is training-free and plug-and-play, enabling any Video-LM to perform fine-grained spatio-temporal grounded reasoning.

\section{Know-Show Benchmark}
\label{sec:dataset}

\begin{figure}[tb]
  \centering
  \includegraphics[width=1.\columnwidth]{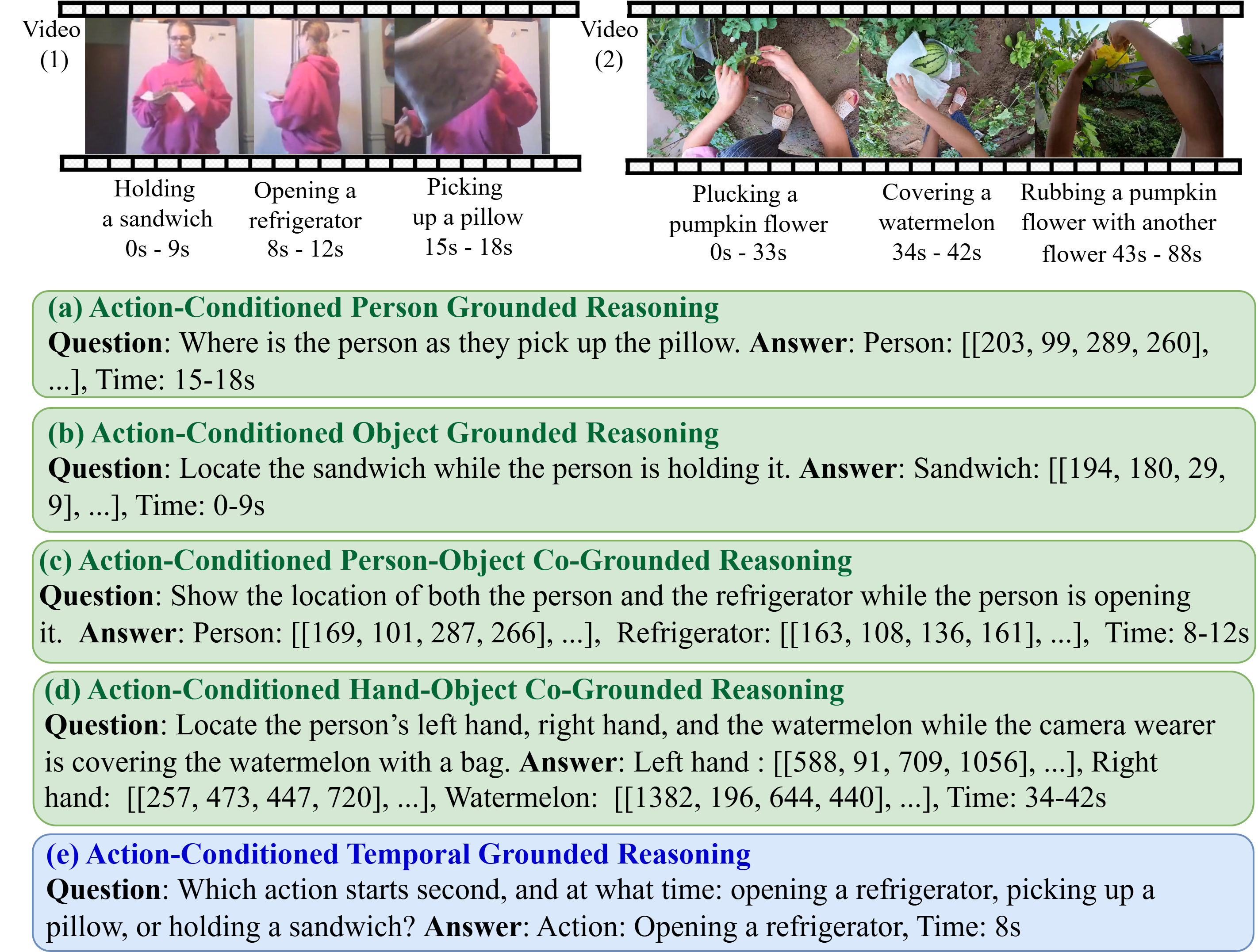}
\caption{
Overview of the test scenarios in the Know-Show Benchmark.
The benchmark consists of two main categories:
(1) Action-Conditioned Spatial Grounded Reasoning, which evaluates a model’s ability to reason about and localize people, objects, and hands within specific action contexts. This category includes four subtypes: (a) person-grounded reasoning, (b) object-grounded reasoning, (c) person–object co-grounded reasoning, and (d) hand–object co-grounded reasoning.
(2) Action-Conditioned Temporal Grounded Reasoning, which assesses the model’s ability to reason about temporal order and to localize actions in time.
Scenarios (a)–(d) are taken from Video~1 (Charades), while scenario (e), corresponding to category (2), is taken from Video~2 (Ego4D).}
  \label{fig:test_Cases}
\end{figure}

% We introduce the Know-Show Benchmark to evaluate the spatio-temporal grounded reasoning capabilities of Video-LMs. It consists of five complementary test scenarios that assess a model’s ability to perform high-level reasoning about actions and their semantics (persons, objects, hands, person-object and hand-object interactions) while grounding these in spatial and temporal structure. Each scenario is built from high-quality human-authored questions drawn from three source datasets.
% A test instance is a tuple ${V, Q, A}$, where $V$ is a video clip, $Q$ is a question requiring joint reasoning and grounding to generate answer $A$. This section outlines the construction of the benchmark from the source datasets and the design and motivation of the five scenarios.
\sloppy
We introduce the Know-Show Benchmark to evaluate the spatio-temporal grounded reasoning abilities of Video-LMs. It contains five complementary test scenarios that probe high-level action understanding and reasoning covering persons, objects, hands, and their interactions, while requiring both spatial and temporal grounding. Each scenario is derived from high-quality, human-authored questions sourced from three datasets. 
A test instance is a tuple $(V, Q, A)$, where $V$ is a video clip, $Q$ is a question requiring both reasoning and grounding, and $A$ is the answer. This section describes how the benchmark is constructed and the motivation behind each test scenario.

\subsection{Source Datasets} 
% \noindent \textbf{Source Datasets.} 
The source datasets were selected to ensure that (1) videos contain human actions with explicit temporal localization, and (2) actions capture semantics involving humans, objects, their interactions, and spatial localization. Based on these criteria, we choose Charades \cite{charades}, Action Genome \cite{ag}, and Ego4D \cite{ego4d}.
%\noindent \textbf{Charades \& Action Genome}. 
Charades provides realistic indoor activities with overlapping and co-occurring actions, enabling evaluation of complex action reasoning. Action Genome augments Charades with spatio-temporal scene graph annotations, including object categories, bounding boxes, and human–object interactions.
%\noindent \textbf{Ego4D}. 
Ego4D is a large egocentric video dataset spanning diverse real-world scenarios. Its explicit hand–object interaction annotations, including bounding boxes and action localization, make it well suited for assessing hand-object co-grounded reasoning scenario explained in Section \ref{sec:test_scenarios}.

\subsection{Benchmark Construction}
%We construct our benchmark from the test splits of the selected datasets. To ensure quality and consistency, two volunteer human annotators were employed for the annotation process. The annotators were first provided with clear instructions on reasoning types associated with each test scenario.  
% We construct the benchmark from the test splits of the selected datasets. To ensure quality and consistency, two trained annotators produced all annotations after receiving clear guidelines on the reasoning types required for each test scenario.
We construct the benchmark from the test splits of the selected datasets, with two trained annotators creating all annotations following detailed guidelines tailored to each reasoning scenario.

\noindent \textbf{Video selection.} 
%The annotators carefully reviewed the videos and discarded those of very low quality (e.g., overly dark or ambiguous clips) where even humans would struggle to interpret the content. Such cases were more frequently observed in Charades. Although Charades is described as clips containing a single person performing multiple actions, we observed a considerable number of clips featuring multiple people. Unlike prior work such as \cite{sugandhika2025situational}, which exclude such cases, we retained these videos and specifically utilized them for the action-conditioned person-grounded reasoning category described in Section \ref{sec:test_scenarios}.
The annotators reviewed all videos and removed those of very low quality (e.g., overly dark or ambiguous clips) that even humans would find difficult to interpret, with such cases occurring more often in Charades. Although Charades is intended to contain a single person performing multiple actions, many clips include multiple people \cite{sugandhika2025situational}. We use them for the action-conditioned person-grounded reasoning category described in Section \ref{sec:test_scenarios}.

\noindent \textbf{Additional annotations.} Since Action Genome provides annotations for only one person, for Charades videos with multiple people, we manually added bounding box annotations for the additional person(s) whose annotations were missing from Action Genome that has only single person annotation. Moreover, although Charades includes action start and end timestamps, we found that some temporal annotations were imprecise or unrealistic. For instance, an action such as “closing a door” typically lasts less than five seconds, yet we found instances labelled as lasting 10–25 seconds.
In such cases, annotators carefully re-examined such clips and refined the temporal boundaries by identifying the precise start and end of each action to reflect their true temporal boundaries.

\noindent \textbf{Action Sequence Selection. } 
When generating test questions in action- conditioned temporal grounded reasoning scenario, we avoided action sequences that could be trivially inferred using common sense knowledge alone. For example, a sequence such as opening the refrigerator → taking milk → closing the refrigerator → drinking milk can often be predicted without visual input by relying on everyday common sense reasoning. 
% Including such cases would reduce the discriminative power of our benchmark.  
Instead, we deliberately selected actions at a finer granularity and less predictable order, ensuring that the correct answer requires genuine video understanding. For example, a sequence such as sitting on a sofa → drinking coffee → opening a refrigerator in a single video cannot be inferred solely from common sense priors. 
In such cases, models must take advantage of the right spatial and temporal embeddings to correctly identify the temporal order and then perform grounding.

\subsection{Know-Show Test Scenarios}
\label{sec:test_scenarios}

We motivate and describe our test scenarios below. Fig \ref{fig:test_Cases} shows an example of each test scenario. 
The benchmark consists of five complementary test scenarios grouped into two categories:
(1) Action-Conditioned Spatial Grounded Reasoning, which evaluates a model’s ability to jointly reason and localize objects, people, hands or their interactions within specific action contexts. This category includes four sub-scenarios in increasing order of compositional complexity: (a) person grounded reasoning, (b) object grounded reasoning, (c) person-object co-grounded reasoning and (d) hand-object co-grounded reasoning.
(2) Action-Conditioned Temporal Grounded Reasoning, which assesses whether the model can reason about the relative temporal order of actions and their absolute localization.

\noindent \textbf{(a) Action-conditioned person grounded reasoning. }
This scenario evaluates whether the model can correctly \emph{link an action to the person performing it and localize that person within the action’s temporal window}, requiring the combined capabilities of action recognition and localization, action–person binding, reasoning and action-conditioned person localization.

\noindent \textbf{(b) Action-conditioned object grounded reasoning. }
This scenario evaluates whether the model can correctly \emph{link an action to the object involved in it and localize that object within the action’s temporal window}. It requires action recognition and localization, action–object binding, reasoning and action-conditioned object localization.

\noindent \textbf{(c) Action-conditioned person-object co-grounded reasoning. }
This scenario evaluates whether the model can correctly \emph{link an action to both the person and the object involved and localize them within the action’s temporal window}, demanding integrated skills in action recognition and localization, action–role binding and reasoning between the interacting entities and action-conditioned localization of the involved person and the object.

\noindent\textbf{(d) Action-conditioned hand–object co-grounded reasoning.}
This scenario focuses on fine-grained contact-level reasoning, evaluating whether the model can correctly \emph{link an action to the specific hand–object interaction and localize both left and the right hands and the object involved in the action context}. It requires precise action recognition and localization, accurate role binding between interacting entities, fine-grained interaction reasoning and action-conditioned spatial grounding of hands and objects.
% within the temporal context of the action.

\noindent \textbf{(e) Action-conditioned temporal grounded reasoning. }
This scenario evaluates whether the model can correctly \emph{link actions to their temporal positions in the video, determining both their relative order and their absolute start/end time}. It requires accurate action recognition, relative temporal order reasoning, and action localization.

\noindent \textbf{Statistics.} Our benchmark includes 2.5K questions, with 500 questions from each of the five categories, constructed from 750 videos drawn from the test sets of the source datasets. Refer Supplementary Section \ref{sup:dataset_stat} for more statistics.

\section{GRAM Framework}
\label{sec:method}

As shown in Fig. \ref{fig:intro_diagram}, despite their impressive progress in various video-language tasks, Video LMs' reasoning remains weakly grounded in spatial and temporal evidence \cite{nextgqa}. 
Although they can recognize objects and capture coarse temporal order, they often fail to explicitly bind the correct humans, objects, and actions to their precise spatial and temporal contexts \cite{nextgqa, yuan2025date}. 
% due to the soft attentional mechanisms of Transformers.
Building on these observations, we propose a training-free, plug-and-play framework, named \textbf{GRAM} (\textbf{G}rounded \textbf{R}easoning \textbf{A}ugmentation \textbf{M}odule), to augment existing Video-LMs with spatio-temporal grounded reasoning capabilities.

\begin{figure}[t]
  \centering
  \includegraphics[width=\linewidth]{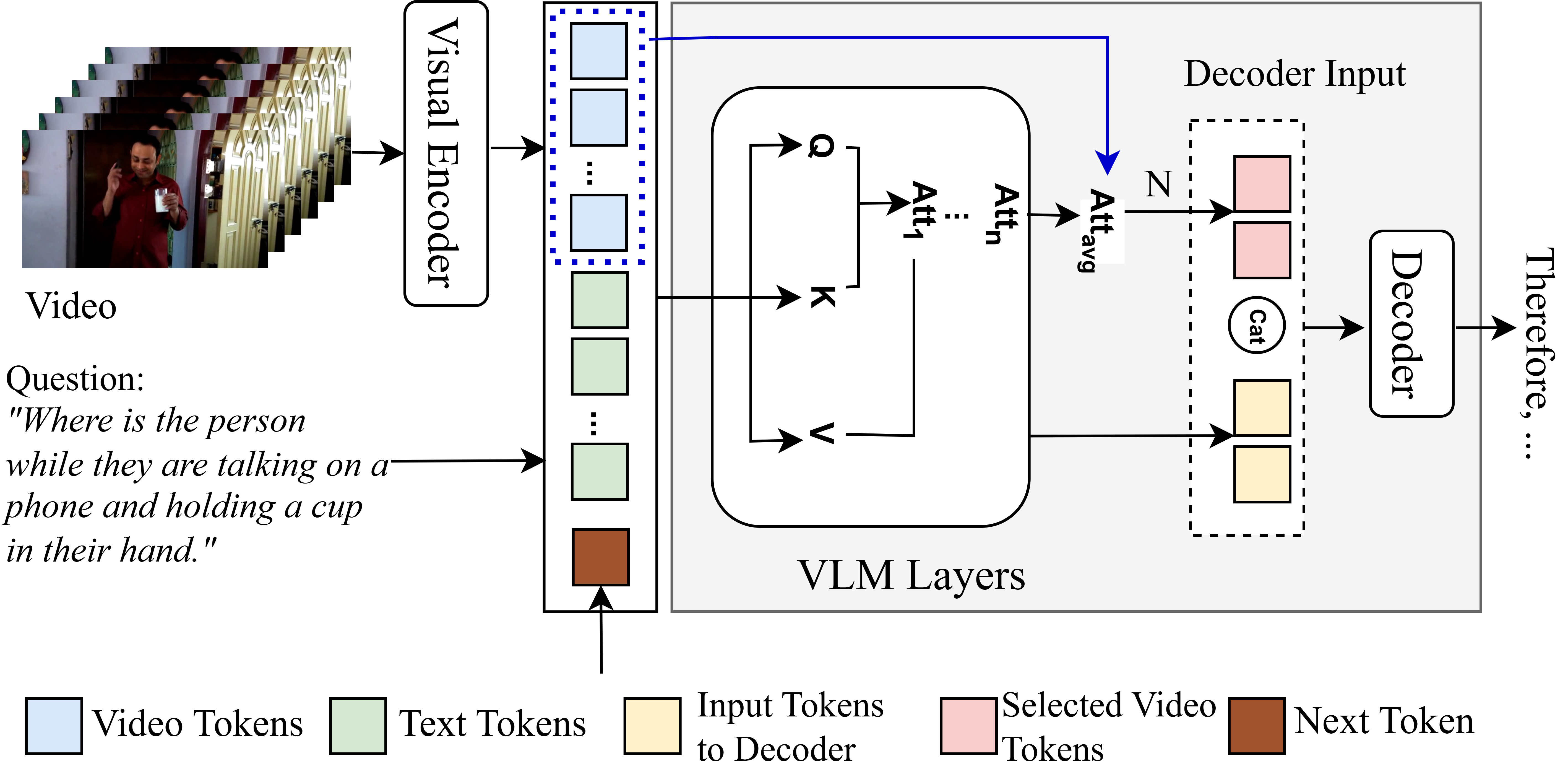}
  \caption{ 
  The inputs consist of a video and a corresponding question. The video is first processed by Vision Encoder to produce video tokens, which are combined with text tokens and previously generated tokens to form the input sequence. During decoding, whenever the model encounters a token that marks the beginning of a new reasoning step, we aggregate the attention of that token across all VLM layers and heads by averaging. From this aggregated attention map, we select the top $N$ most attended video tokens. These selected tokens are then concatenated with the Decoder’s input embeddings and fed into the Decoder to generate the next token. This iterative process ensures that each reasoning step is spatially and temporally grounded in the video, facilitating \textbf{Spatio-Temporal Grounded Reasoning}.
  }
  \label{fig:architecture}
\end{figure}

% \begin{figure}[t]
%   \centering
%   \resizebox{0.7\linewidth}{!}{%
%     \includegraphics{images/architecture_v3.png}
%   }
%   \caption{Illustration of the decoding process in our \textbf{GRAM} plugin.}
%   \label{fig:architecture}
% \end{figure}

\noindent \textbf{Preliminaries.}
A video-LM consists of a Visual Encoder and generative language model. 
First, the video $V$ is sampled into a sequence of frames $\{f_1, f_2,..., f_t \}$ and is passed through the Visual Encoder to convert them into input video tokens, i.e., the token representations produced from raw video input.
These tokens are inserted into the textual instruction at designated video placeholder positions.
The combined sequence of text and video tokens is then processed by the model to produce either a direct answer or a step-wise reasoning sequence, i.e., $rs_1, rs_2, ...$, followed by a final answer especially when prompted with cues such as \emph{“Think step by step”} \cite{zheng2023ddcot} as
$(rs_1, rs_2, \ldots, \text{Answer}) = \mathrm{VLM}(\text{Video},\allowbreak\, \text{Instruction},\allowbreak\, \text{Prompt})$

% 
% 
% \begin{equation}
% \resizebox{0.9\linewidth}{!}{$
% (rs_1, rs_2, \ldots, \text{Answer})
% = \mathrm{VLM}(\text{Video},\, \text{Instruction},\, \text{Prompt})
% $}
% \label{eq:vlm_outputs}
% \end{equation}
% 
% 
In our setup, we focus on this step-wise reasoning behaviour, aiming to encourage the model to explicitly ground each reasoning step on the relevant visual evidence in the video before producing the final answer. 
% This allows us to assess whether the model’s final prediction and the reasoning leading to it is truly grounded in the observed video.
% spatio-temporal content rather than inferred from global video context or linguistic priors.

\subsection{Spatio-Temporal Grounded Reasoning}

In this work, we introduce a baseline formulation for fine-grained, spatio-temporal grounded reasoning, ensuring that each reasoning step generated by the model is explicitly supported by the most relevant evidence in the video. Our approach builds on the framework proposed in \cite{gao2025interleaved} for image-grounded reasoning and extends it to the video domain.
Our goal is to guide the model to reason based on what it truely sees rather than on the global context or language priors. In this work, we take Qwen2.5-VL as our baseline Video-LM.
We modify its decoding process as shown in Fig. \ref{fig:architecture}. At each decoding step, the model predicts the next token $next_{tok}$ conditioned on the tokens generated so far. 
Specifically, whenever a token $\mathcal{S}$ indicating the beginning of a new reasoning step (e.g., after a full stop) is encountered, it identifies the $N$ video tokens most attended by $\mathcal{S}$. 
To achieve this, we compute the average attention from $\mathcal{S}$ across every layer and head of the VLM, and then select the top top $N$ most attended video tokens from the input video token set (blue dashed box in Fig. \ref{fig:architecture}).
These selected video tokens $V_{stok}$ (pink boxes in Fig. \ref{fig:architecture}), are concatenated with the input tokens $input_{tok}$ at the embedding level 
where, $input_{tok}$ denotes the Decoder input used for the autoregressive text generation process.
The next token's decoding step is then formulated as,
% This is done by taking the average attention map of $\mathcal{S}$ across all layers and heads of the VLM, denoted as $\mathcal{A}$, and computing a dot-product similarity between $\mathcal{A}$ and the video tokens $V_{tok}$ as,
% \begin{equation}
% Sim = \mathcal{A} \cdot V_{tok}.
% \label{eq:attention_similarity}
% \end{equation}
% \noindent The top $N$ most relevant video tokens are then selected.
% These selected video tokens $V_{stok}$ (pink boxes in Fig. \ref{fig:architecture}), are concatenated with the input tokens $input_{tok}$ at the embedding level
% where, $input_{tok}$ denotes the Decoder input used for the autoregressive text generation process.
% The next token's decoding step is then formulated as,
\begin{eqnarray}
next_{tok} &=& \mathrm{Decoder}\big( 
    \mathrm{Cat}(input_{tok}^{emb},\, V_{stok}^{emb})
\big)
\label{eq:next_token_generation}
\end{eqnarray}
\noindent where $input_{tok}^{emb}$ and $V_{stok}^{emb}$ denote the embedding of the input tokens and selected video tokens respectively, and $\mathrm{Cat}(.)$ represents feature concatenation. 
This procedure ensures that the model’s next-token prediction is guided by spatially and temporally localized visual evidence.
% , promoting fine-grained spatio-temporal grounded reasoning.
Through this formulation, each generated reasoning step is explicitly grounded to the spatio-temporal regions in the video that support it, providing an alignment between language-based reasoning and visual evidence, which is an essential step towards closing the gap between reasoning and grounding in Video-LMs.

\subsection{Explicit Timestamp Tokens}

For spatio-temporal grounded reasoning, a model must not only identify the visual evidence supporting its reasoning but also ensure that this visual evidence is aligned with the correct temporal moments.
Although Qwen2.5-VL claims to incorporate absolute time awareness, our results show that most current models, including Qwen2.5-VL achieve less than half of human-level performance in absolute temporal localization.
Specifically, Qwen2.5-VL employs Multimodal Rotary Position Embedding (MRoPE) \cite{qwen}, a unified mechanism that extends rotary embeddings to support 1D, 2D, and 3D modalities.
Text tokens are assigned 1D positional indices, while visual tokens receive 3D indices corresponding to time (T), height (H), and width (W), allowing the model to apply 3D RoPE to visual tokens and 1D RoPE to textual tokens.
When processing videos as sequences of frames, the temporal position ID is incremented for each frame, while the height and width components follow the same assignment scheme as in static images, thereby embedding the absolute temporal information of video frames.

However, empirically, we find that such implicit positional encoding alone is insufficient for achieving accurate absolute temporal localization, even though it performs reasonably well on relative temporal reasoning.
To address this limitation, we augment Qwen with explicit timestamp tokens (e.g., \texttt{<0.0s>},'' \texttt{<5.1s>},'' ``\texttt{<8.3s>}''), which are interleaved with video frame tokens in the input sequence.
These timestamp tokens serve as temporal anchors, granting the model direct access to absolute time rather than relying solely on implicit temporal embeddings.
This explicit temporal representation encourages the model to develop a semantically consistent notion of absolute time, leading to better temporal alignment and more precise grounding of reasoning steps, without requiring any model retraining.

\section{Results}
\label{sec:results}

\noindent \textbf{Evaluated Video-LMs.}
We evaluate multiple open video-LMs: Qwen2.5 VL 7B \cite{qwen}, VideoChat2 \cite{li2024mvbench}
% VideoLLaVA \cite{videollava}, 
LLaVA-OneVision \cite{llavaonevision}, Ovis2.5-9B \cite{ovis}, MiniCPM-V 4.5 \cite{minicpm}, 
Video-UTR-7B \cite{yu2025unhackable},
LongVA \cite{zhang2024long}, VideoR1 \cite{feng2025video}; closed models GPT-4o \cite{gpto} and Gemini 3.1 Pro \cite{gemini}. 
We additionally conduct a human evaluation on a Know-Show subset consisting of 250 randomly selected samples, with 50 samples drawn from each evaluation category.

\subsection{Evaluation Metrics}
% \noindent \textbf{Evaluation Metrics.}
We evaluate both Action-Conditioned Spatio-Temporal Grounded Reasoning (SGR) and Action-Conditioned Temporal Grounded Reasoning (TGR) using Accuracy as the primary metric.

\noindent \textbf{Evaluating SGR.}
For each sample $i$, let $\hat{B}_i^{(x)}$ denote the predicted bounding box for entity type $x$, 
and $B_i^{(x)} = \{B_{i,1}^{(x)}, B_{i,2}^{(x)}, \dots, B_{i,K_i}^{(x)}\}$ represent the set of all ground-truth boxes 
for the same entity within the annotated action duration $[t_i^s, t_i^e]$ where $t_i^s$ and $t_i^e$ refer to the ground truth start and end time stamps.  
Here, $K_i$ denotes the number of annotated frames (and thus ground-truth boxes) for entity type $x$ within the action duration.  
The Intersection-over-Union (IoU) $x\text{IoU}$ for entity type $x$ and temporal accuracy $T_{Acc}$ is defined as:

\begin{equation}
x\text{IoU}_i = 
\max_{B \in B_i^{(x)}} 
\frac{|\hat{B}_i^{(x)} \cap B|}{|\hat{B}_i^{(x)} \cup B|}; \quad T_{Acc} = \hat{t}_i \in [t_i^s, t_i^e],
\end{equation}
where the IoU is computed between the predicted box $\hat{B}_i^{(x)}$ and every ground-truth box in $B \in B_i^{(x)}$,  
and the maximum IoU value across all ground-truth boxes is taken as the final $x\text{IoU}_i$. 
Here, $x \in \{\text{P}, \text{O}, \text{LH}, \text{RH}\}$, where $\text{P}$, $\text{O}$, $\text{LH}$, and $\text{RH}$ correspond to the person, object, left hand, and right hand, respectively.  
For example, $PIoU$ is referring to Person-IoU. $\hat{t}_i$ is the predicted timestamp.
Spatial grounded reasoning accuracy $Acc^{SGR}$ is defined as,
\begin{equation}
Acc^{SGR} = x\text{IoU}_i \ge \tau \land T_{Acc},
\end{equation}
\noindent where $\tau = 0.25$ is the IoU threshold.
% \blue{ where $\tau \in \{0.25, 0.5\}$ is the IoU threshold.}

\noindent \textbf{Evaluating TGR.}
For each sample $i$, let $\hat{a}_i$ and $a_i$ denote the predicted and ground-truth action labels, respectively. 
Let $t_i$ denote the ground-truth temporal annotations, where $t_i$ corresponds to either the start time or the end time of the action, depending on the question.  
A prediction is considered correct within a temporal window $\delta \in \{2, 4, 6\}$ seconds if, 

% \begin{equation}
% A_{Acc} =(\hat{a}_i = a_i); \quad T@\delta =|\hat{t}_i - t_i| \le \delta; \quad Acc^{TGR} =A_{Acc} \land T@\delta,
% \end{equation}
\begin{equation}
\resizebox{\linewidth}{!}{$
A_{Acc} =(\hat{a}_i = a_i); \quad T@\delta =|\hat{t}_i - t_i| \le \delta; \quad Acc^{TGR} =A_{Acc} \land T@\delta
$}
\end{equation}
where $A_{Acc}$, $T@\delta$ and $Acc^{TGR}$ denote the action accuracy, time stamp accuracy and temporal grounded reasoning accuracy.
We additionally report the Mean Absolute Deviation $MAD$ between the predicted and ground-truth start/end time to show how closely the predicted action onset aligns with the actual ground-truth time.

We apply GRAM to Qwen 2.5 7B model. Implementation details and prompts are included in the supplementary Section~\ref{sup:implementation}.

% \paragraph{Implementation Details}

% \red{We draw inspiration from dual-process theories of reasoning in cognitive science, which distinguish between fast, heuristic-driven “System 1” thinking and slower, deliberative “System 2” reasoning. Single-step prompting resembles System 1-style inference, requiring models to jointly infer actions, participants, and their spatial-temporal grounding in a single forward pass, often relying on implicit correlations. In contrast, our multi-step evaluation protocol imposes a structured, coarse-to-fine reasoning process that isolates action understanding, temporal localization, and spatial grounding. This decomposition mirrors System 2-like behavior by increasing inference-time computation and reducing shortcut heuristics. Evaluating both settings enables us to probe whether current Video-LMs possess intrinsic grounded reasoning ability or benefit from explicit reasoning scaffolds.}

\subsection{Main results}
\label{sec:main_results}

\noindent We evaluate existing Video-LMs and our GRAM baseline across the spatial and temporal grounded reasoning categories in Table \ref{tab:SGR} and \ref{tab:TGR} respectively. The bold and underline font shows the best and the second best result respectively. PGR, OGR, POCoGR, HOCoGR and TGR refer to action-conditioned person, object, person-object, hand-object and temporal grounded reasoning scenarios.

\begin{table*}[tb]

\centering
\resizebox{1.0\textwidth}{!}{
\begin{tabular}{|l|
                c |c| >{\cellcolor{gray!15}}c|
                c |c| >{\cellcolor{gray!15}}c|
                c |c |c| >{\cellcolor{gray!15}}c|
                c |c |c |c| >{\cellcolor{gray!15}}c|}
                % c| c| c|>{\cellcolor{gray!15}}c |}

\hline
% \multirow{2}{*}{Model} & 
% \multicolumn{3}{c|}{Person Grnd. Reasoning (PGR)} &
% \multicolumn{3}{c|}{Object Grnd. Reasoning (OGR)} &
% \multicolumn{4}{c|}{Person–Object Co-Grnd. Reasoning (POCoGR)} &
% \multicolumn{5}{c|}{Hand–Object Co-Grnd. Reasoning (HOCoGR)} \\

\multirow{2}{*}{Model} & 
\multicolumn{3}{c|}{PGR} &
\multicolumn{3}{c|}{OGR} &
\multicolumn{4}{c|}{POCoGR} &
\multicolumn{5}{c|}{HOCoGR} \\

\cline{2-16}

& $T_{Acc}$ & $PIoU$ & $Acc^{SGR}$
& $T_{Acc}$ & $OIoU$ & $Acc^{SGR}$
& $T_{Acc}$ & $PIoU$ & $OIoU$ & $Acc^{SGR}$
& $T_{Acc}$ & $LHIoU$ & $RHIoU$ & $OIoU$ & $Acc^{SGR}$ \\
% & $MAD$ & $A_{Acc}$ & $T@2$ & $Acc^{TGR}$ \\

\hline
% Qwen2.5 VL 3B (2024) & 30.1 & 23.5 & 8.4 & 33.6 & 5.6 & 1.0 & 33.5 & 39.4 & 8.5 & 0.8 & 2.0 & 0.8 & 0.8 & 0.5 & 0.0 \\
Qwen (2024) & 19.8 & 29.4 & 6.3 & 24.0 & 8.2 & 2.4 & 26.8 & 43.8 & 10.8 & 2.6 & 2.5 & 3.5 & 2.0 & 1.5 & 0.0 \\
% \red{VideoLLaVA} & 0.6 & 62.8 & 0.2 & 1.2 & 23.8 & 0.2 & 0.9 & 0.0 & 0.0 & 0.0 & 3.2 & 0.8 & 0.8 & 0 & 0.0 \\
VideoChat2 (2024) & 20.4 & 11.7 & 5.3 & 7.6 & 3.4 & 3.4 & 12.3 & 18.4 & 12.5 & 2.2 & 3.2 & 0.0 & 0.0 & 0.0 & 0.0 \\
LLaVA-OneVision (2024) & 25.1 & 37.8 & 9.0 & 31.2 & 14.0 & 4.6 & 24.0 & 35.2 & 48.0 & 5.2 & 3.7 & 0.8 & 0.5 & 0.8 & 0.0 \\
Ovis (2024) & 48.7 & 34.3 & 16.1 & 49.6 & 10.6 & 5.8 & 51.6 & 34.8 & 9.4 & 4.5 & 3.5 & 0.3 & 0.3 & 0.3 & 0.0 \\
LongVA (2024) & 19.6 & 24.6 & 7.6 & 24.4 & 15.8 & 5.8 & 24.0 & 35.7 & 45.6 & 5.1 & 4.9 & 0.0 & 0.0 & 0.0 & 0.0 \\
MiniCPM (2025) & 45.4 & 8.2 & 2.7 & 40.4 & 1.4 & 0.4 & 50.4 & 17.1 & 4.0 & 0.4 & 3.2 & 1.2 & 0.8 & 0.3 & 0.0 \\
Video-UTR (2025) & 16.6 & 2.4 & 1.8 & 15.8 & 0.4 & 0.1 & 19.3 & 45.4 & 1.97 & 0.2 & 0.0 & 0.0 & 0.0 & 0.0 & 0.0 \\
VideoR1 (2025) & 38.9 & 69.5 & \underline{20.8} & 38.6 & 18.8 & 6.9 & 51.5 & 78.1 & 15.1 & 6.6 & 0.1 & 0.0 & 0.0 & 0.0 & 0.0 \\

\hdashline[1pt/2pt]
GPT-4o (2024) & 38.4 & 36.6 & 10.6 & 35.8 & 13.4 & 4.4 & 20.6 & 10.8 & 5.3 & 3.7 & 1.3 & 0.0 & 0.0 & 0.0 & 0.0 \\
Gemini (2025) & 79.3 & 21.5 & 20.1 & 78.0 & 17.5 & \underline{13.8} & 79.8 & 26.1 & 16.8 & \underline{14.8} & 7.2 &	10.0 &	8.5	& 19.5	& \textbf{0.3} \\

\hdashline[1pt/2pt]
% \textbf{GRAM (Ours)} 448 & 25.6 & 81.4 & \textbf{20.9} & 50.4 & 46.0 & \textbf{27.8} & 46.7 & 66.7 & 31.8 & \textbf{12.7} & 4.2 & 1.0 & 0.5 & 0.0 & 0.0 \\
% \textbf{GRAM (Ours)} 512 & 33.4 & 80.1 & 26.6 & 50.2 & 37.0 & 21.2 & 39.3 & 80.7 & 73.1 & 28.0 & 3.5 & 5.8 & 0.0 & 0.0 & 0.0 \\
GRAM (Ours) & 33.4 & 80.1 & \textbf{26.6} & 50.2 & 37.0 & \textbf{21.2} & 39.3 & 80.7 & 73.1 & \textbf{28.0} & 3.5 & 5.8 & 0.0 & 0.0 & 0.0 \\

\hdashline[1pt/2pt]
Human & 84.0 & 92.0 & 82.0 & 89.3 & 100.0 & 94.7 & 81.3 & 94.7 & 88.0 & 74.7 & 100 & 83.6 & 85.7 & 91.1 & 79.5 \\
\hline
\end{tabular}
}
\caption{Results on Action-Conditioned Spatial-Grounded Reasoning on Know-Show}
\label{tab:SGR}
\end{table*}

\begin{table*}[tb]

\centering
\resizebox{\textwidth}{!}{
\begin{tabular}{|l|c|c|c|>{\columncolor[gray]{0.9}}c|c|>{\columncolor[gray]{0.9}}c|c|>{\columncolor[gray]{0.9}}c||l|c|c|c|>{\columncolor[gray]{0.9}}c|c|>{\columncolor[gray]{0.9}}c|c|>{\columncolor[gray]{0.9}}c|}
\hline

Model & 
$A_{Acc}$ & 
$MAD$ &
\multicolumn{2}{c|}{2 seconds} &
\multicolumn{2}{c|}{4 seconds} &
\multicolumn{2}{c|}{6 seconds} &

Model & 
$A_{Acc}$ & 
$MAD$ &
\multicolumn{2}{c|}{2 seconds} &
\multicolumn{2}{c|}{4 seconds} &
\multicolumn{2}{c|}{6 seconds} \\

\cline{4-9} 
\cline{13-18}
& & &
$T@2$ & $Acc^{TGR}$ &
$T@4$ & $Acc^{TGR}$ &
$T@6$ & $Acc^{TGR}$ &
& & &
$T@2$ & $Acc^{TGR}$ &
$T@4$ & $Acc^{TGR}$ &
$T@6$ & $Acc^{TGR}$ \\
\hline

% Qwen2.5 VL 3B (2024) & 52.5 & 5.8 & 19.4 & 11.8 & 45.2 & 28.7 & 59.8 & 35.6 &
% MiniCPM (2025) & 47.9 & 58.3 & 22.2 & 14.5 & 39.6 & 24.7 & 52.3 & 31.8 \\

% Qwen2.5 VL 7B (2024) & 56.0 & 6.6 & 28.1 & 15.7 & 48.8 & \underline{29.7} & 58.6 & 34.6 &
% Video-UTR (2025)  & 39.1 & 23.7 & 27.0 & 13.2 & 27.0 & 13.1 & 52.1 & 23.7 \\

% VideoChat2 (2024) & 98.5 & 43.1 & 19.0 & \underline{18.8} & 28.2 & 27.7 & 34.4 & 33.8 &
% VideoR1 (2025) & 48.6 & 18.0 & 27.0 & 11.2 & 41.8 & 20.1 & 52.1 & \underline{37.0} \\

% LLaVA-OneVision (2024) & 48.9 & 25.1 & 16.8 & 10.0 & 39.5 & 23.9 & 50.0 & 29.3 &
% GPT-4o (2024) & 68.0 & 14.9 & 44.1 & \textbf{34.0} & 65.3 & \textbf{47.1} & 75.7 & \textbf{54.3} \\

% \cdashline{10-18}[1pt/2pt]
% Ovis (2024) & 53.6 & 23.9 & 17.8 & 12.0 & 37.8 & 24.2 & 50.9 & 32.5 &
% GRAM (Ours) & 48.7 & 22.1 & 20.5 & 11.8 & 42.7 & 26.9 & 56.2 & 33.2 \\

% \cdashline{10-18}[1pt/2pt]
% Long-VA (2024) & 35.6 & 26.4 & 16.5 & 7.5 & 32.7 & 17.2 & 40.3 & 20.1  &
% Human & 94.4 & 0.7 & 94.4 & 94.4 & 95.8 & 94.4 & 100 & 94.4 \\

Qwen (2024) & 56.0 & 6.6 & 28.1 & 15.7 & 48.8 & 29.7 & 58.6 & 34.6 &
Video-UTR (2025) & 39.1 & 23.7 & 27.0 & 13.2 & 27.0 & 13.1 & 52.1 & 23.7 \\

VideoChat2 (2024) & 98.5 & 43.1 & 19.0 & 18.8 & 28.2 & 27.7 & 34.4 & 33.8 &
VideoR1 (2025) & 48.6 & 18.0 & 27.0 & 11.2 & 41.8 & 20.1 & 52.1 & 37.0 \\

\cdashline{10-18}[1pt/2pt]
LLaVA-OneVision (2024) & 48.9 & 25.1 & 16.8 & 10.0 & 39.5 & 23.9 & 50.0 & 29.3 &
GPT-4o (2024) & 68.0 & 14.9 & 44.1 & \underline{34.0} & 65.3 & \underline{47.1} & 75.7 & \underline{54.3} \\

Ovis (2024) & 53.6 & 23.9 & 17.8 & 12.0 & 37.8 & 24.2 & 50.9 & 32.5 &
Gemini (2025) & 70.4 & 8.2 & 76.6 & \textbf{60.1} & 85.1 & \textbf{64.5} & 90.2 & \textbf{66.0} \\

\cdashline{10-18}[1pt/2pt]
Long-VA (2024) & 35.6 & 26.4 & 16.5 & 7.5 & 32.7 & 17.2 & 40.3 & 20.1 &
GRAM & 48.7 & 22.1 & 20.5 & 11.8 & 42.7 & 26.9 & 56.2 & 33.2 \\

\cdashline{10-18}[1pt/2pt]
MiniCPM (2025) & 47.9 & 58.3 & 22.2 & 14.5 & 39.6 & 24.7 & 52.3 & 31.8 &
Human & 94.4 & 0.7 & 94.4 & 94.4 & 95.8 & 94.4 & 100 & 94.4 \\

% 
% VideoLLaVA & 29.8 & 53.3 & 7.1 & 4.0 & 14.1 & 5.7 & 19.7 & 7.5 \\

\hline
\end{tabular}
}
\caption{Results on Action-Conditioned Temporal Grounded Reasoning on Know-Show}
\label{tab:TGR}
\end{table*}

% \vspace{0.2em}
\noindent\textbf{Action-Conditioned Spatial-Grounded Reasoning.}
Across majority of the Video-LMs, we observe a consistent performance hierarchy: \small PGR $>$ OGR $>$ POCoGR $>$ HOCoGR. 
PGR is relatively easier, likely because humans are highly salient visual entities and large-scale pre-training strongly biases models toward detecting and representing people. 
Although several models achieve better IoU@0.25 (e.g., GPT-4o and VideoR1), the corresponding $Acc^{SGR}$ remains substantially lower ($\sim$20\%), indicating that models often localize a salient person but fail to bind the correct person to the queried action within its temporal window.
% GRAM improves substantially, reaching 81\% (IoU@0.25) and 46\% (IoU@0.5), outperforming the both open and closed models with 21\% and 12\%, overall performance respectively.
% However, performance remains limited since the task requires not merely detecting a person, but correctly binding the person to the specific action within the relevant temporal window. 
% VideoR1 and GRAM achive highest performance.
% As localization thresholds become stricter, performance declines across all models; however, GRAM exhibits comparatively more graceful degradation (e.g., relative to VideoR1). 
% (e.g., while GRAM and VideoR1 perform similarly at IoU@0.25, at IoU@0.5 GRAM drops by 8\%, compared to an 11\% drop for VideoR1.)
% 
% also say in extdended analysis, objects getting higher IoU without temporal accuracy means, can detected objects, but not under action constraint
Gemini, VideoR1 and GRAM achieve higher performance in this category, as reflected by their $Acc^{SGR}$ scores.
% As localization thresholds become stricter, performance declines across all models; however, GRAM exhibits comparatively more graceful degradation. For example, while GRAM and VideoR1 perform similarly at IoU@0.25, at IoU@0.5 GRAM drops by 8\%, compared to an 11\% drop for VideoR1; detailed results at IoU@0.5 are provided in Supplementary Section~\ref{full_SGR}.
% 
OGR is more challenging, as objects are often smaller, occluded, or less visually salient, and object–action associations may be weaker and more ambiguous as multiple objects and instances appear. 
Most models achieve $\sim$15\% at IoU@0.25. GRAM outperforms with overall Accuracy $Acc^{SGR}$ of 21\% (IoU@0.25).
% and 10\% (IoU@0.5).
% 
The difficulty increases further in POCoGR, where the model must jointly localize two entities, preserve correct role assignment (who interacts with what), and ground both within the action’s temporal context, thereby exposing limited compositional reasoning capacity.
While models can often recognize the action and localize the person or object individually at levels comparable to earlier tasks, their performance further drops when these requirements must be satisfied simultaneously. This suggests that current Video-LMs tend to treat actions, people, and objects in isolation rather than reasoning over their localized spatio-temporal interactions.
Despite the overall difficulty, GRAM achieves $Acc^{SGR}$ 28\% at IoU@0.25, still low in absolute terms, but higher than all models. 
Furthermore, we observe a consistent gap in $Acc^{SGR}$ between IoU@0.25 and IoU@0.5 across these reasoning categories and this indicates that models rely on coarse, region-level localization rather than boundary-accurate grounding; detailed results at IoU@0.5 are provided in Supplementary Section~\ref{sup:full_SGR}.
% —indicating that better alignment between reasoning and localized visual evidence improves person–object association and co-grounding.
% 
% The most challenging scenario is HOCoGR, where performance drops sharply across all models except in Gemini. 
The most challenging scenario is HOCoGR, where performance declines notably across all models, although Gemini shows slightly better performance than the others. We believe this may be due to Gemini’s proprietary training data, architectural and optimization strategies, which are not publicly disclosed.
This setting requires fine-grained contact reasoning including left–right hand distinction and object detection, and precise spatial localization of fine-grained interactions - capabilities that seems to be poorly supported by current Video-LMs, which often operate on coarse visual tokens and lack explicit modeling of fine-grain interaction-level constraints.
Being an egocentric setting, temporal accuracy across all models also drops substantially compared to third-person performance in other categories \cite{nunez2022egocentric}. We believe that these failures likely stem from limited egocentric and hand–object interaction data in pretraining \cite{lin2022egocentric}, the small size and rapid motion of hands and objects, and encoders’ dependence on coarse scene-level cues rather than fine spatial detail required for physical interaction modeling \cite{schroder2017hand, bao2024handsonvlm}.
% 
% \red{At IoU@0.25, only Gemini attains non-negligible performance 3.5\%, likely benefiting from broader proprietary training exposure.} 

% \vspace{0.2em}
\noindent\textbf{Action-Conditioned Temporal Grounded Reasoning.}
Compared to spatial grounding, Video-LMs perform relatively better on temporal grounded reasoning, with models such as Gemini and GPT-4o reaching $\sim$30-60\% $Acc^{TGR}$. However, Table~\ref{tab:TGR} reveals a clear gap between relative temporal order reasoning and temporal localization: although $A_{Acc}$ is high, $T@\delta$ drops significantly under stricter thresholds, indicating limited precision in identifying \emph{when} an action occurs. The consistent improvement of $T@\delta$ with larger tolerance windows (2 $\rightarrow$ 4 $\rightarrow$ 6s), together with high MAD values, suggests that models capture coarse temporal regions rather than accurate event boundaries. Overall, current Video-LMs exhibit relative temporal awareness but struggle with absolute fine-grained temporal grounding. Qualitative examples are provided in Supplementary section~\ref{sup:qualitative_results}.

\noindent \textbf{Zero-shot vs CoT inference.}
The prompts used in Table~\ref{tab:SGR} and Table~\ref{tab:TGR} follow CoT style, as detailed in Supplementary ~\ref{sup:implementation} (Fig.~\ref{fig:GRAM_prompts}). To evaluate the effect of CoT prompting, we further assess Qwen and GPT-4o under a zero-shot setting without explicit CoT cues (Supplementary \ref{sup:multi_step}, Table~\ref{tab:ablation_cot}). We observe no meaningful performance difference between the two configurations;
e.g., Qwen achieves 6.3/5.5 and GPT-4o 10.6/11.0 on PGR under CoT/zero-shot prompting, respectively.
In addition, we explore a multi-step prompting strategy that decomposes each query into sequential sub-questions. However, this approach also does not produce consistent or substantial gains (Supplementary \ref{sup:multi_step}, Table~\ref{tab:ablation_step_by_step}).
Overall, these results indicate that prompt engineering alone is insufficient to improve grounded reasoning in Video-LMs. While structured prompts may encourage more organized outputs, they do not resolve underlying spatio-temporal representation and grounding limitations, indicating the need for architectural and representation-level solutions rather than prompt-based guidance alone.

% \subsection{Ablation study of GRAM}
\noindent\textbf{Ablation study of GRAM.}
We evaluate the contributions of each component in the GRAM framework in Table~\ref{tab:ablation}. Specifically, we compare four variants: (A) base Qwen 2.5 7B model, (B) spatio-Temporal Grounded Reasoning (ST-GR),
where each reasoning step produced by the model is explicitly grounded by relevant spatio-temporal evidence in the video,
% where the model incorporates the video embeddings most similar to the current token's attention context when predicting the next token, 
(C) Explicit Timestamp Tokens (ETT) where we interleave absolute time tokens with frame embeddings in the input sequence, and (D) the full GRAM model, which combines both (B) and (C).
First, we find that adding grounded reasoning (variant B) provides clear improvements, especially in spatial localization, indicating the benefit of grounding next-token prediction in the relevant visual evidence.
% Second, adding dense ordinal temporal encoding with explicit time-stamps (variant C) further improves performance over implicit temporal encoding alone, highlighting that modelling time as a structured signal is crucial for precise grounded reasoning. Combining both enhancements (variant D) yields the best results overall, confirming that spatial-temporal grounding and temporal alignment are complementary to acheive spatio-temporal grounded reasoning.
Second, Explicit Timestamp Tokens (variant C) further improves performance beyond the implicit temporal cues in (A), yielding strong gains in spatial localization and boosting overall accuracy.
Finally, when grounded reasoning is augmented with explicit timestamp tokens (variant D), the model attains the best overall performance, demonstrating that explicit temporal encoding helps better align the visual evidence with the correct action moments in time.
Additional ablations on the effect of the number of input video frames
% and resolution are 
is provided in Supplementary section \ref{sup:ablation_num_frames}.

\begin{table*}[tb]
\centering
\resizebox{\textwidth}{!}{

\begin{tabular}{|l|
                c |c| >{\cellcolor{gray!15}}c|
                c |c| >{\cellcolor{gray!15}}c|
                c |c |c| >{\cellcolor{gray!15}}c|
                c |c |c |c| >{\cellcolor{gray!15}}c|
                c| c| c|>{\cellcolor{gray!15}}c |}

\hline
\multirow{2}{*}{Model} & 
\multicolumn{3}{c|}{PGR} &
\multicolumn{3}{c|}{OGR} &
\multicolumn{4}{c|}{POCoGR} &
\multicolumn{5}{c|}{HOCoGR} &
\multicolumn{4}{c|}{TGR} \\

\cline{2-20}

& $T_{Acc}$ & $PIoU$ & $Acc^{SGR}$
& $T_{Acc}$ & $OIoU$ & $Acc^{SGR}$
& $T_{Acc}$ & $PIoU$ & $OIoU$ & $Acc^{SGR}$
& $T_{Acc}$ & $LHIoU$ & $RHIoU$ & $OIoU$ & $Acc^{SGR}$
& $A_{Acc}$ & $MAD$ & $T@2$ & $Acc^{TGR}$ \\

\hline

(A) Qwen2.5 VL 7B  & 19.8 & 29.4 & 6.3 & 24.0 & 8.2 & 2.4 & 26.8 & 43.8 & 10.8 & 2.6 & 2.5 & 3.5 & 2.0 & 1.5 & 0.0 & 56.0 & 6.6 & 28.1 & \underline{15.7} \\
(B) Qwen2.5 VL 7B  + ST-GR   & 16.6 & 46.9 & 9.7 & 11.2 & 10.4 & 2.2 & 23.0 & 35.3 & 10.2 & 2.1 & 2.2 & 0.5 & 0.0 & 1.3 & 0.0 & 32.1 & 47.8 & 22.1 & 8.7 \\
(C) Qwen2.5 VL 7B  + ETT & 29.2 & 48.1 & \underline{18.2} & 16.2 & 34.2 & \underline{9.2} & 25.7 & 50.7 & 20.2 & \underline{6.6} & 2.2 & 0.5 & 0.0 & 0.3 & 0.0 & 16.6 & 36.8 & 20.7 & 4.3 \\
% (D) GRAM (A+B+C) & 25.6 & 81.4 & \textbf{20.9} & 50.4 & 46.0 & \textbf{27.8} & 46.7 & 66.7 & 31.8 & \textbf{12.7} & 4.2 & 1.0 & 0.5 & 0.0 & 0.0 & 48.7 & 22.1 & 36.1 & \textbf{23.7} \\
(D) GRAM (A+B+C) & 33.4 & 80.1 & \textbf{26.6} & 50.2 & 37.0 & \textbf{21.2} & 39.3 & 80.7 & 73.1 & \textbf{28.0} & 3.5 & 5.8 & 0.1 & 0.0 & 0.0 & 48.7 & 22.1 & 36.1 & \textbf{23.7} \\

\hline
\end{tabular}
}
\caption{Ablation study of GRAM on Know-Show benchmark. 
Results are reported using IoU@0.25 for all Action-Conditioned Spatial Grounded Reasoning tasks.}
\label{tab:ablation}
\end{table*}

\noindent\textbf{Outlook.}
Our benchmark clearly demonstrates that recognition alone, reasoning alone or spatial/temporal localization alone are each insufficient for robust spatio-temporal grounded reasoning.
While current Video-LMs consistently struggle across scenarios, human performance exceeds 75\% across all metrics, exposing a significant gap and substantial room for improvement in spatio-temporal grounded reasoning capabilities of the Video-LMs.
Despite being training-free, GRAM achieves consistent gains by explicitly aligning intermediate reasoning steps with localized spatial and temporal evidence, directly mitigating a key representation bottleneck i.e., weak alignment of reasoning with grounded evidence; however, results remain far from human-level performance. Looking forward, meaningful progress in this direction will likely require architectural mechanisms that,
(i) enhance grounded reasoning, 
(ii) explicit relational modelling of action-role binding with grounding (who interacts with what, when, and where), 
(iii) incorporate fine-grained interaction dynamics, 
(iv) fine-grained spatial supervision for objects specially under blur and occluded settings,
(v) pre-training tailored to egocentric perspectives, and 
(vi) integrate explicit temporal supervision  than relying solely on implicit temporal embeddings.

\section{Conclusion}
\label{sec:conclusion}

In this work, we revisited a fundamental question: Do today’s Video-LMs truly perform grounded reasoning, do they show what they know and know what they show? Our findings indicate that they struggle with this task. Models often recognize actions without grounding them in the correct person, object, hand, or moment, or correctly localize entities while inferring the wrong action. This misalignment between reasoning and evidence is exactly what our Know-Show benchmark exposes.
To narrow this gap, we introduced GRAM, a simple training-free baseline that strengthens the link between predictions and video cues, yielding consistent improvements across spatial-temporal grounded reasoning tasks.
% Looking forward, advancing grounded video reasoning will require Video-LMs with stronger spatio-temporal grounded reasoning, finer modelling of objects and hands, explicit relational reasoning, and better temporal supervision, key capabilities essential for models that truly show what they know and know what they show.
% \newline
Looking ahead, progress will require Video-LMs with stronger spatio-temporal grounded reasoning mechanisms, finer modelling of objects and hands, explicit and grounded relational reasoning, and robust temporal supervision, the capabilities essential for models that both show what they know and know what they show, ultimately enabling reliable and faithful grounded reasoning.

\noindent \textbf{Acknowledgment} This research/project is supported by the National Research Foundation, Singapore, under its NRF Fellowship (Award\# NRF-NRFF14-2022-0001) and funding allocation to B.F. by A*STAR under its SERC Central Research Fund (CRF).

{
    \small
    \bibliographystyle{ieeenat_fullname}
    \bibliography{main}
}

% WARNING: do not forget to delete the supplementary pages from your submission 
\clearpage
\setcounter{page}{1}
\maketitlesupplementary

% \noindent \textbf{Know-Show:Benchmarking Video-Language Models on Spatio-Temporal Grounded Reasoning: Supplementary Material}

% \begin{center}
% \textbf{Know-Show: Benchmarking Video-Language Models on Spatio-Temporal Grounded Reasoning: Supplementary Material}
% \end{center}

\noindent In this supplementary material, we present:
\begin{enumerate}
    \item Further details about Know-Show benchmark including benchmark construction guidelines, statistics and human evaluation setup in section \ref{sup:dataset_stat})
    \item Implementation details in section \ref{sup:implementation}
    \item Performance on spatial-grounded reasoning at higher IoU thresholds in section \ref{sup:full_SGR}
    \item Discussion on Zero-shot, Chain of Thought (CoT) and multi-step prompting in Section \ref{sup:multi_step}
    
    \item Ablation studies of GRAM on input resolution and number of video frames in Section \ref{sup:ablations}
    \item A visual ablation study of GRAM for spatio-temporal grounded reasoning in section \ref{sup:gram_reasoning}
    % \item Ablation study of GRAM on number of frames in section \ref{sup:ablation_num_frames}
    % \item Human evaluation setup in section \ref{sup:human_eval_setup}
    \item Qualitative results of model predictions in section \ref{sup:qualitative_results}
\end{enumerate}

\noindent In addition to this document, we also \textbf{release a 10\% subset of our Know-Show benchmark} in \textbf{Annex A}. 
We will also release the codebase for our GRAM plug-in upon acceptance of the paper.
% 4. Limitations (Appendix D).}

\

% ###################################################################################

\section{Further Details about Know-Show Benchmark}
\label{sup:dataset_stat}

\subsection{Guidelines for Creating Questions}
Annotators first reviewed each video to assess its suitability for annotation. Videos of very low quality (e.g., overly dark, visually ambiguous, or difficult to interpret even for humans) were filtered out and excluded from the dataset.
Prior to the annotation process, annotators were provided with detailed explanations and examples for each question type to ensure a clear understanding of the required reasoning categories. After reviewing the guidelines, annotators watched the selected videos and generated questions based on the reasoning types.
% , ensuring that the questions were grounded in the visual content of the video.
Bounding box annotations were primarily sourced directly from the original datasets.
% to maintain consistency with their spatial annotations. 
In cases where the source datasets did not provide the necessary annotations, annotators manually added bounding boxes. For example, in multi-person interaction videos, additional bounding boxes were provided to annotate the required person instance when such annotations were not available in the original dataset. Specially, Charades and Action Genome provide bounding box annotations for only a single person instance.

% \noindent \textbf{Dataset statistics.}
\subsection{Benchmark Statistics}
Source videos for the reasoning scenarios involving action-conditioned person, object, person–object, and temporal grounded reasoning were obtained from the Charades and Action Genome datasets, while the action-conditioned hand–object co-grounded reasoning scenario was sourced from the Ego4D dataset.
Each question category contains 500 questions. 
The dataset contains a total of 675 videos, of which 625 come from Charades and 50 from Ego4D. On average, each video contains approximately 1.05 ± 0.2 people in Charades and 1.4 ± 1.1 people in Ego4D, resulting in an overall average of 1.08 ± 0.37 people per video. The number of annotated objects per video is 5.1 ± 1.7 in Charades and 3.2 ± 0.7 in Ego4D, with an overall average of 4.96 ± 1.72 objects per video.
In terms of actions, Charades videos contain 10.4 ± 4.1 actions per video, while Ego4D videos contain 1.9 ± 2.6 actions per video, leading to an overall average of 9.77 ± 4.58 actions per video. The average action duration is 12 ± 8.9 seconds for Charades and 35 ± 64.5 seconds for Ego4D, with an overall mean action length of 13.70 ± 20.44 seconds.
The occlusion rate in Charades is 20.3 ± 15.9\%, while this statistic is not available for Ego4D. Regarding video duration, Charades videos have an average length of 28.9 ± 7.6 seconds, whereas Ego4D videos are significantly longer, averaging 266.4 ± 74.1 seconds. Overall, the combined dataset has an average video length of 47.97 ± 70.77 seconds.
Finally, the annotation density, measured as the number of bounding boxes per question, is 21.5 ± 14.7 for Charades and 10.3 ± 36.6 for Ego4D, resulting in an overall average of 25.7 ± 27.3 bounding boxes per question.

% Further dataset statistics are presented in Table \ref{tab:dataset_stat}. Annotation density was calculated using the ground-truth bounding boxes for all reasoning types across both datasets.

% \begin{table}[h]
% % \caption{Dataset statistics for Charades and Ego4D used in Know-Show.}
% % \label{tab:dataset_stat}
% \centering
% \resizebox{0.8\textwidth}{!}{
% \begin{tabular}{|l|c|c|c|}
% \hline
% \textbf{Statistic} & \textbf{Charades} & \textbf{Ego4D} & \textbf{Overall} \\
% \hline
% Number of videos & 625 & 50 & 675 \\
% Number of people per video & 1.05$\pm$0.2 & 1.4$\pm$1.1 & 1.08$\pm$0.37 \\
% Number of objects per video & 5.1$\pm$1.7 & 3.2$\pm$0.7 & 4.96$\pm$1.72 \\
% Number of actions per video & 10.4$\pm$4.1 & 1.9$\pm$2.6 & 9.77$\pm$4.58 \\
% Action length (seconds) & 12$\pm$8.9 & 35$\pm$64.5 & 13.70$\pm$20.44 \\
% Occlusion rate (\%) & 20.3$\pm$15.9 & N/A & N/A \\
% Video length (seconds) & 28.9$\pm$7.6 & 266.4$\pm$74.1 & 47.97$\pm$70.77 \\
% Annotation density (bounding boxes per question) & 21.5\pm$14.7$  & 10.3\pm$36.6$ & 25.7\pm$27.3$ \\ 
% % Clip length (s) & N/A & 7.9 / 0.1 & N/A \\
% \hline
% \end{tabular}
% }
% % \caption{Dataset statistics for Charades and Ego4D used in Know-Show.}
% \end{table}

% \noindent \textbf{Annotator Agreement.} 
\subsection{Annotator Agreement}
To evaluate annotation reliability, we conducted an inter-annotator agreement study on a subset of the dataset. 
Two annotators independently re-annotated samples originally labeled by the other annotator. 
The temporal annotation deviations, measured as mean ($\mu$) and standard deviation ($\sigma$), were low, indicating strong consistency between annotators: start time ($\mu=0.4$, $\sigma=0.5$) and end time ($\mu=0.7$, $\sigma=0.6$). 
Additionally, both video–question alignment and question–action alignment achieved 100\% agreement. 
% Bounding box annotations were directly adopted from the source datasets, ensuring consistency with their original spatial annotations.

% % \noindent \textbf{Annotators’ Background.}
% \subsection{Annotator Agreement}
% Both annotators have academic backgrounds in Computer Science and reported at least a fluent level of proficiency in English.

\subsection{Human Evaluation}
\label{sup:human_eval_setup}

We conducted human evaluations to establish a human performance baseline on the Know-Show benchmark.
% The same subset used for evaluating the closed Video-LMs was employed for this study.
Four volunteers, comprising postgraduate and undergraduate students with a background in Computer Vision, were recruited to answer questions spanning all reasoning categories. Their responses were then compared directly with model outputs using identical evaluation criteria.
To ensure consistent evaluation, all questions were distributed via Google Forms with relavant instructions. For each question, evaluators were asked to first read the question and then watch the associated video. 
We provided pre-extracted video frames.
Participants selected the most relevant frame and provided the associated bounding box and timestamp using the VGG Image Annotator tool \cite{tool}.

% ###################################################################################

\section{Implementation Details}
\label{sup:implementation}

\begin{figure}[t]
    \centering
    \begin{subfigure}[]{1.0\linewidth}
        \centering
        \includegraphics[width=1.0\linewidth]{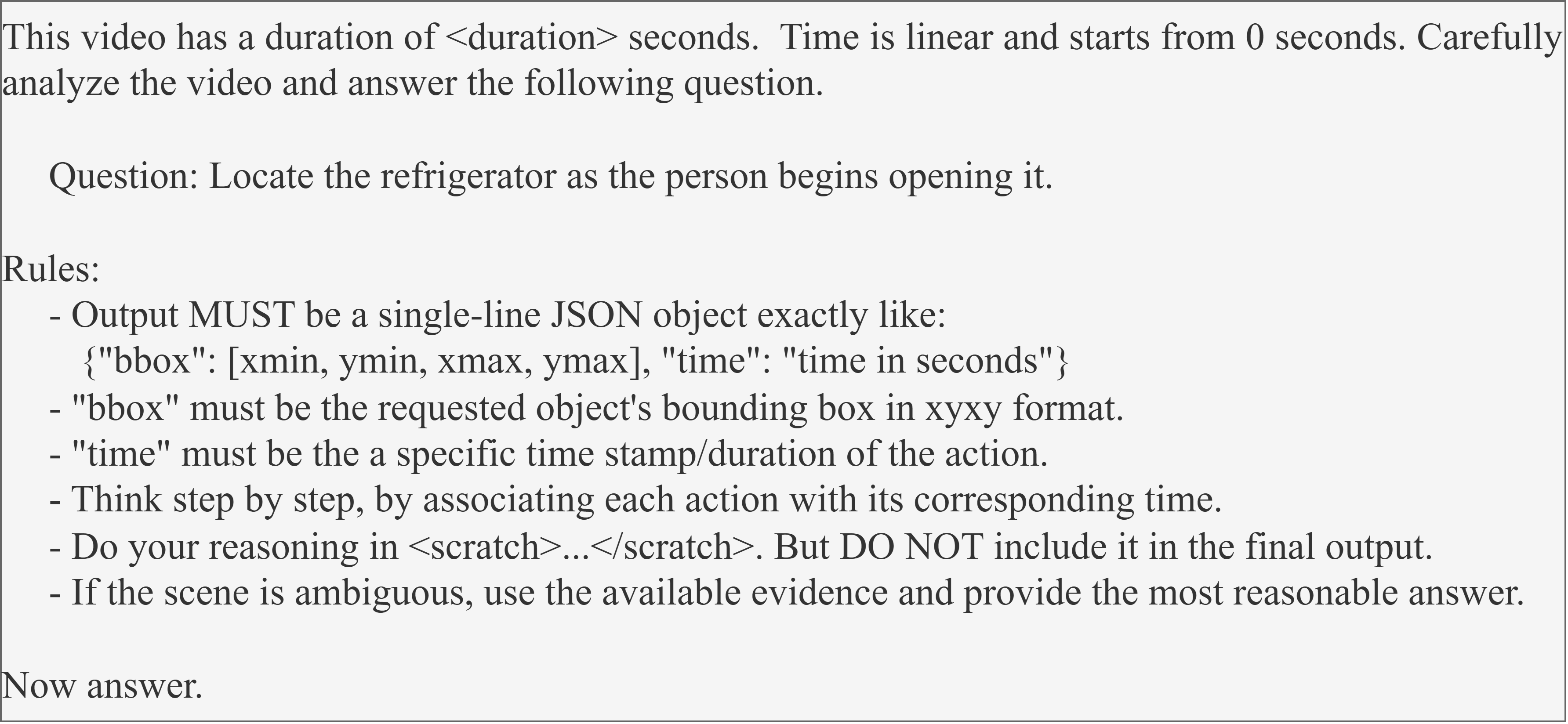}
        \subcaption{Prompt used for Action Conditioned Spatial Grounded Reasoning scenarios.}
        \label{}
    \end{subfigure}
    \hfill
    \begin{subfigure}[]{1.0\linewidth}
        \centering
        \includegraphics[width=1.0\linewidth]{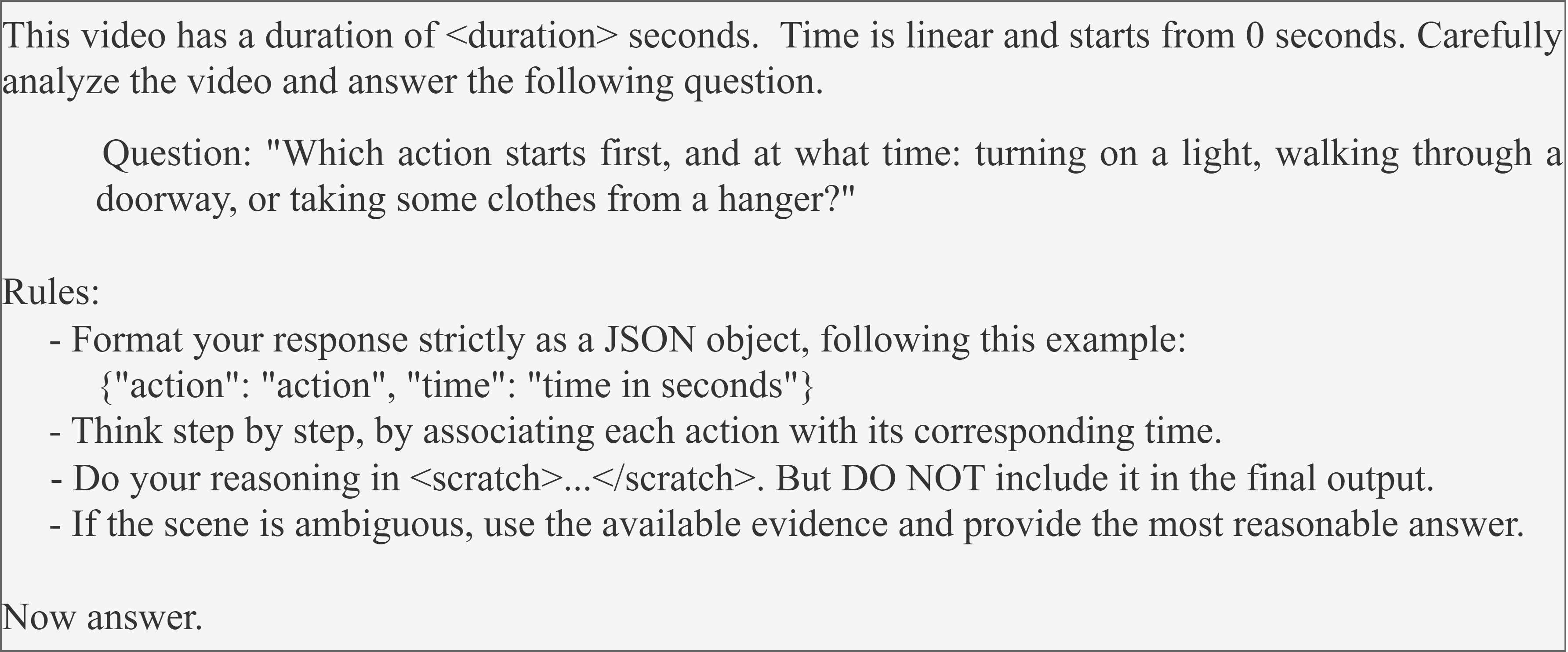}
        \subcaption{Prompt used for Action Conditioned Temporal Grounded Reasoning scenario.}
        \label{}
    \end{subfigure}
    \caption{Prompts used to evaluate the Video-LMs.}
    \label{fig:GRAM_prompts}
\end{figure}

We evaluated all Video-LMs using the official evaluation scripts provided by HuggingFace or their official web pages. All experiments were conducted on a single NVIDIA A100 SXM4 80GB GPU. 
For our GRAM plug-in, we set the number of most-attended video tokens selected to $N=64$, following empirical observations reported in \cite{gao2025interleaved}. We used the prompts shown in Fig.~\ref{fig:GRAM_prompts} to evaluate all models reported in Tables \ref{tab:SGR}, \ref{tab:TGR}, and \ref{tab:ablation}.

\noindent \textbf{Output Parsing.}
To control for output format inconsistencies across models, we implemented model-specific parsing functions to enforce a unified bounding box format. Predicted bounding boxes were normalized to the original frame resolution when necessary (e.g., LongVA and Ovis etc. produced normalized bboxes). We further manually verified the parsed outputs and visualized the predicted boxes overlaid on the corresponding frames, ensuring that the results were not affected by parsing issues or coordinate misalignment.
During evaluation, we observed that when prompted to predict “time,” particularly in Action-Conditioned Spatial Grounded Reasoning scenarios, models sometimes produced either a single timestamp or a temporal duration. For duration-based predictions, we used the temporal midpoint of the predicted interval as the final timestamp for evaluation. In contrast, for Temporal Grounded Reasoning scenarios, models consistently produced a single timestamp.

% \red{Discuss about output parsing: 
% We controlled for format mismatch by using model-specific parsing functions, enforcing a unified box format and normalizing boxes to the original frame resolution if required. We manually verified parsed outputs and visualized boxes overlaid on frames, confirming that the low localization scores are not due to parsing or coordinate errors.}

% ##################################################################################
\section{Performance on Spatial-Grounded Reasoning at Higher IoU Thresholds}
\label{sup:full_SGR}

% \begin{itemize}
%     \item \blue{all models performance drops at IoU=0.5, But GRAM and Geminis seems perofming in a similar manner although GRAM perfoms best at IoU@0.25. This may be because Gemini secures higher temporal accuracy (e.g. above 70\% in PGR, OGR, POCoGR) }
%     % \item \blue{GRAM secures relatively higher IoU specially in PGR, OGR and POCoGR}
%     \item \blue{Gemini is performing well in temporal accuracy as they did in TGR scenarios. this suggest that Gemini has both relative and absolute temporal understanding, what theu lack is to link spatial gronding withr elavant temporal context.}
%     \item \blue{Temporal acc drops from thrid person to first person e.g. gemini, from 70\% from frist person to 7.3\% in first person. Reasons codul be as discussed in section \ref{sec:main_results}, limited exposure of the video LMs into ego-cntric pretraining. another reason could be the length of the Ego-cnetric videos could be 4.4 seconds and first person videos on average have 30 seconds. And annotationd density tcould also be a reason for the lower performance. in third person setting has more dense annotations in Table \ref{tab:dataset_stat} annotation density per question for Action Genome is 21 where as for Ego4D is 10}
%     \item \blue {There is a huge gap between the models and humans and a room for future improvement. This suggest, human hav ethis capability of ground ed reasoning naturlllay, but video-LMs lack this capability.}
%     \item \blue{}
%     \item \blue{}
% \end{itemize}

All models exhibit a noticeable performance decline when the IoU threshold increases from 0.25 to 0.5, indicating that precise spatial grounded reasoning remains challenging. 
% While GRAM achieves the best performance at IoU@0.25, it performs comparably to Gemini at IoU@0.5. This similarity may be attributed to Gemini’s strong temporal accuracy, which exceeds 70\% in all PGR, OGR, and POCoGR, enabling more consistent temporal alignment during grounded reasoning. 
GRAM and Gemini achieve the highest performance across PGR, OGR, and POCoGR at IoU@0.5, with both models showing comparable results. Overall, Gemini demonstrates strong temporal reasoning capabilities even in Spatial Grounded Reasoning tasks, consistent with its performance in the Temporal Grounded Reasoning scenario. This suggests that the model is effective at capturing both relative and absolute temporal relationships in videos.
However, Gemini appears to struggle with precise spatial localization, indicating difficulty in effectively linking spatial grounding with the relevant temporal context.
% However, its weaker spatial localization indicates difficulty in effectively linking spatial grounding with the relevant temporal context. 
Furthermore, temporal accuracy drops across most of the models significantly when moving from third-person (PGR, OGR, POCoGR) to first-person (HOCoGR) videos. 
% For instance, Gemini’s performance decreases from around 70\% in third-person settings to approximately 7.3\% in first-person scenario. 
This drop may be due to several factors: limited exposure of Video-LMs to egocentric data during pretraining as discussed in Section \ref{sec:main_results}, longer video durations in the egocentric dataset (approximately 366 seconds on average compared to 30 seconds in third-person videos, and lower annotation density (annotation density per question is higher in Action Genome $\sim$21 compared to Ego4D $\sim$10), which may reduce the supervision available for grounding in egocentric setting. 
Finally, a substantial performance gap remains between models and humans across all reasoning types, suggesting that while humans naturally possess spatio-temporal grounded reasoning capabilities, current Video-LMs still struggle to reason about actions and semantics while simultaneously grounding their inferences in the appropriate spatial and temporal evidence.

\begin{table*}[h]

\centering
\resizebox{\textwidth}{!}{
\begin{tabular}{
|l|
c|c|>{\cellcolor{gray!15}}c|c|>{\cellcolor{gray!15}}c| % 5 cols
c|c|>{\cellcolor{gray!15}}c|c|>{\cellcolor{gray!15}}c| % 5 cols
c|c|c|>{\cellcolor{gray!15}}c|c|c|>{\cellcolor{gray!15}}c| % 7 cols
c|c|c|c|>{\cellcolor{gray!15}}c| % 5 cols
}
\hline
Model & \multicolumn{5}{c|}{PGR} & 
\multicolumn{5}{c|}{OGR} & 
\multicolumn{7}{c|}{POCoGR} & 
% \multicolumn{5}{c|}{Hand-Object Co-Grounded Reasoning} \\
\multicolumn{5}{c|}{HOCoGR} \\

\cline{2-23}

& $T_{Acc}$ & \multicolumn{2}{c|}{$IoU$ $>=$ 0.25} & \multicolumn{2}{c|}{$IoU$ $>=$ 0.5} & 
$T_{Acc}$ & \multicolumn{2}{c|}{$IoU$ $>=$ 0.25} & \multicolumn{2}{c|}{$IoU$ $>=$ 0.5} & 
$T_{Acc}$ & \multicolumn{3}{c|}{$IoU$ $>=$ 0.25} & \multicolumn{3}{c|}{$IoU$ $>=$ 0.5} &
$T_{Acc}$ & \multicolumn{4}{c|}{$IoU$ $>=$ 0.25} \\

\cline{3-6} \cline{8-11} \cline{13-18} \cline{20-23}

 &  & $PIoU$ & $Acc^{SGR}$ & $PIoU$ & $Acc^{SGR}$ & 
    & $OIoU$ & $Acc^{SGR}$ & $OIoU$ & $Acc^{SGR}$ &
    & $PIoU$ & $OIoU$ & $Acc^{SGR}$  & $PIoU$ & $OIoU$ & $Acc^{SGR}$ &
    & $LHIoU$ & $RHIoU$ & $OIoU$ & $Acc^{SGR}$ \\
\hline

% Qwen2.5 VL 3B & 30.1 & 23.5 & 8.4 & 6.9 & 2.4 & 33.6 & 5.6 & 1.0 & 0.8 & 0.0 & 33.5 & 39.4 & 8.5 & 0.8 & 10.8 & 1.8 & 0.2 & 2.0 & 0.8 & 0.8 & 0.5 & 0.0 \\
Qwen & 19.8 & 29.4 & 6.3 & 8.0 & 2.0 & 24.0 & 8.2 & 2.4 & 2.2 & 1.0 & 26.8 & 43.8 & 10.8 & 2.6 & 11.0 & 2.6 & 0.4 & 2.5 & 3.5 & 2.0 & 1.5 & 0.0 \\
% VideoLLaVA & 0.6 & 62.8 & 0.2 & 24.9 & 0.0 & 1.2 & 23.8 & 0.2 & 6.6 & 0.2 & 0.9 & 0.0 & 0.0 & 0.0 & 0.0 & 0.0 & 0.0 & 3.2 & 0.8 & 0.8 & 0.0 & 0.0 \\
VideoChat2 & 20.4 & 11.7 & 5.3 & 4.1 & 1.8 & 7.6 & 3.4 & 3.4 & 1.6 & 1.2 & 12.3 & 18.4 & 12.5 & 2.2 & 4.0 & 3.7 & 0.1 & 3.2 & 0.0 & 0.0 & 0.0 & 0.0 \\
LLaVA-OneVision & 25.1 & 37.8 & 9.0 & 6.3 & 1.0 & 31.2 & 14.0 & 4.6 & 1.4 & 0.8 & 24.0 & 35.2 & 48.0 & 5.2 & 9.7 & 20.5 & 0.4 & 3.7 & 0.8 & 0.5 & 0.8 & 0.0 \\
Ovis & 48.7 & 34.3 & 16.1 & 13.3 & 4.9 & 49.6 & 10.6 & 5.8 & 1.6 & 1.2 & 51.6 & 34.8 & 9.4 & 4.5 & 12.8 & 1.6 & 0.6 & 3.5 & 0.3 & 0.3 & 0.3 & 0.0 \\
MiniCPM & 45.4 & 8.2 & 2.7 & 1.6 & 0.6 & 40.4 & 1.4 & 0.4 & 0.2 & 0.0 & 50.4 & 17.1 & 4.0 & 0.4 & 2.6 & 0.9 & 0.0 & 3.2 & 1.2 & 0.8 & 0.3 & 0.0 \\
Video-UTR & 16.6 & 33.3 & 1.8 & 0.2 & 0.0 & 15.8 & 0.4 & 0.1 & 0.0 & 0.0 & 19.3 & 45.4 & 2.0 & 0.0 & 8.1 & 0.2 & 0.0 & 0.0 & 0.0 & 0.0 & 0.0 & 0.0 \\
Long-VA & 19.6 & 24.6 & 7.6 & 2.2 & 0.8 & 24.4 & 15.8 & 5.8 & 3.8 & 1.8 & 24.0 & 35.7 & 45.6 & 5.1 & 7.0 & 21.8 & 1.0 & 4.9 & 0.0 & 0.0 & 0.0 & 0.0 \\
VideoR1 & 38.9 & 69.5 & \underline{20.8} & 25.2 & 9.6 & 38.6 & 18.8 & 6.6 & 3.6 & 1.8 & 51.5 & 78.1 & 15.1 & 6.6 & 34.9 & 4.6 & 0.7 & 0.1 & 0.0 & 0.0 & 0.0 & 0.0 \\

\hdashline[1pt/2pt]
GPT-4o & 38.4 & 36.6 & 10.6 & 16.6 & 4.9 & 35.8 & 13.4 & 4.4 & 2.8 & 0.6 & 20.6 & 10.8 & 5.3 & 3.7 & 7.0 & 1.3 & 0.9 & 1.3 & 0.0 & 0.0 & 0.0 & 0.0 \\
% Gemini 2.5 Pro & 72.9 & 5.1 & 3.9 & 1.8 & 1.4 & 73.0 & 0.6 & 0.4 & 0.2 & 0.1 & 52.0 & 8.5 & 2.4 & 0.6 & 0.0 & 0.4 & 0.0 & 35.5 & 12.8 & 9.5 & 30.3 & 1.0 \\
Gemini & 79.3 & 21.5 & 20.1 & 7.9 & \underline{14.2} & 78.0 & 17.5 & \underline{13.8} & 9.0 & \underline{7.1} & 79.8 & 26.1 & 16.8 & \underline{14.8} & 24.0 & 12.6 & \underline{10.0} & 7.2 & 10.0 &	8.5	& 19.5	& \textbf{0.3} \\

\hdashline[1pt/2pt]
% GRAM 448 & 25.6 & 81.4 & 20.9 & 46.0 & 12.3 & 50.4 & 46.0 & 27.8 & 18.0 & 10.8 & 46.7 & 66.7 & 31.8 & 12.7 & 27.4 & 10.3 & 2.0 & 4.2 & 1.0 & 0.5 & 0.0 & 0.0 \\
% Human & 84.0 & 92.0 & 82.0 & 86.0 & 76.0 & 89.3 & 100.0 & 94.7 & 96.0 & 89.3 & 81.3 & 94.7 & 88.0 & 74.7 & 94.7 & 76.0 & 70.0 & x & x & x & x & x \\
GRAM & 33.4 & 80.1 & \textbf{26.6} & 40.9 & \textbf{15.3} & 50.2 & 37.0 & \textbf{21.2} & 12.1 & \textbf{7.2} & 39.3 & 80.7 & 73.1 & \textbf{28.0} & 37.6 & 45.2 & \textbf{10.2} & 3.5 & 5.6 & 0.0 & 0.0 & 0.0 \\

\hdashline[1pt/2pt]
Human & 84.0 & 92.0 & 82.0 & 86.0 & 76.0 & 89.3 & 100.0 & 94.7 & 96.0 & 89.3 & 81.3 & 94.7 & 88.0 & 74.7 & 94.7 & 76.0 & 70.0 & 100 & 83.6 & 85.7 & 91.1  & 79.5 \\

\hline
\end{tabular}
}
\caption{Extended Results on Action-Conditioned Spatial-Grounded Reasoning test scenarios on Know-Show benchmark.}
\label{tab:SGR_extended}
\end{table*}

% ###############################################################################################################    
\section{Evaluation of Zero-Shot, Chain-of-Thought, and Multi-Step Reasoning Strategies}
\label{sup:multi_step}

% \red{ToDo add (1) detailed explanations, (2)step-by-step prompts, (3) reasoning traces}

The prompts used in Table~\ref{tab:SGR}, ~\ref{tab:TGR} and \ref{tab:ablation} follow a Chain-of-Thought (CoT) prompting style. To examine the effect of CoT prompting, we additionally evaluate Qwen2.5 VL 7B and GPT-4o under a zero-shot inference setting without explicit CoT instructions (i.e., phrases “Think step by step...”, “Do your reasoning in...”, and “If the scene is ambiguous...”). The results of this comparison are reported in Table~\ref{tab:ablation_cot}. Overall, we observe no significant or meaningful performance difference between the CoT and zero-shot prompting configurations.

\begin{table*}[h]

\centering
\resizebox{\textwidth}{!}{

\begin{tabular}{|l|
                c |c| >{\cellcolor{gray!15}}c|
                c |c| >{\cellcolor{gray!15}}c|
                c |c |c| >{\cellcolor{gray!15}}c|
                c |c |c |c| >{\cellcolor{gray!15}}c|
                c| c| c|>{\cellcolor{gray!15}}c |}

\hline
\multirow{2}{*}{Model} & 
\multicolumn{3}{c|}{PGR} &
\multicolumn{3}{c|}{OGR} &
\multicolumn{4}{c|}{POCoGR} &
\multicolumn{5}{c|}{HOCoGR} &
\multicolumn{4}{c|}{TGR} \\

\cline{2-20}

& $T_{Acc}$ & $PIoU$ & $Acc^{SGR}$
& $T_{Acc}$ & $OIoU$ & $Acc^{SGR}$
& $T_{Acc}$ & $PIoU$ & $OIoU$ & $Acc^{SGR}$
& $T_{Acc}$ & $LHIoU$ & $RHIoU$ & $OIoU$ & $Acc^{SGR}$
& $A_{Acc}$ & $MAD$ & $T@2$ & $Acc^{TGR}$ \\

\hline

Qwen zero-shot  
& 17.1 & 33.6 & 5.5 
& 20.8 & 7.8 & 1.4 
& 23.8 & 43.0 & 10.2 & 2.2 
& 2.7 & 2.4 & 0.3 & 0.9 & 0.0 
& 50.5 & 6.4 & 26.9 & 16.9 \\

Qwen CoT 
& 19.8 & 29.4 & 6.3 
& 24.0 & 5.6 & 2.2 
& 26.8 & 42.4 & 9.1 & 2.0 
& 2.5 & 3.5 & 2.0 & 1.5 & 0.0 
& 55.8 & 6.6 & 19.4 & 10.9 \\

GPT-4o zero-shot & 34.4 & 29.0 & 11.0 
& 39.2 & 10.0 & 2.6 
& 47.3 & 19.3 & 5.9 & 2.9 
& 1.2 & 0.0 & 0.0 & 0.0 & 0.0 
& 63.7 & 15.9 & 40.6 & 28.9 \\

GPT-4o CoT  
& 38.4 & 38.4 & 10.6 
& 35.8 & 13.4 & 4.4 
& 20.6 & 10.8 & 5.3 & 3.7 
& 1.3 & 0.0 & 0.0 & 0.0 & 0.0 
& 68.0 & 14.9 & 44.1 & 34.0 \\

\hline
\end{tabular}
}
\caption{CoT style prompting vs zero-shot inference in Qwen2.5 VL 7B and GPT4o models.}
\label{tab:ablation_cot}
\end{table*}

% (1) “Think step by step ...”,
% (2) “Do your reasoning in \textless scratch \textgreater...\textless /scratch \textgreater ...”, and
% (3) “If the scene is ambiguous ...”.

\begin{figure}[h]
  \centering
  \includegraphics[width=\linewidth]{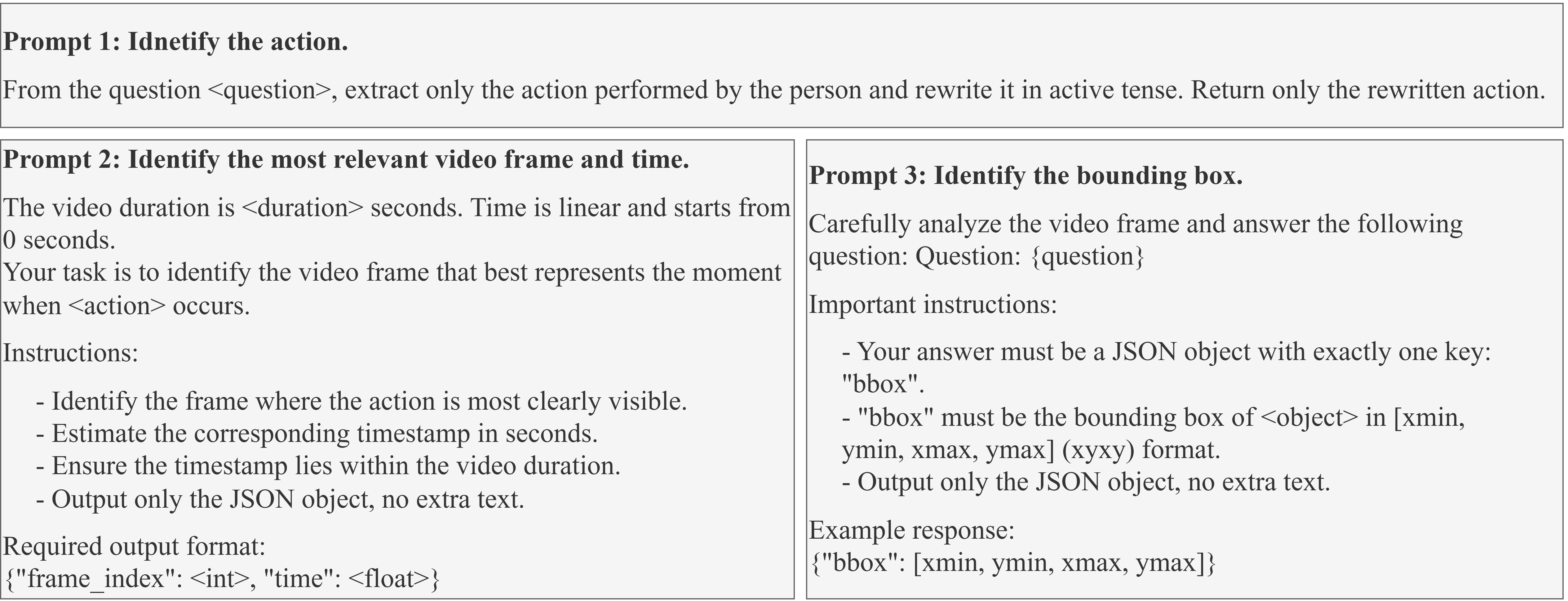}
  \caption{Prompts used for multi-step evaluation.}
  \label{fig:multi_step_prompt}
\end{figure}

We also conduct a multi-step evaluation in which each question is decomposed into a sequence of intermediate sub-questions, as illustrated in Fig.~\ref{fig:multi_step_prompt}. The results for Qwen are reported in Table~\ref{tab:ablation_step_by_step}. However, this strategy also does not produce noticeable or consistent improvements over the standard prompting setup. In fact, multi-step approach significantly increases the average inference time per question to approximately 21 seconds, whereas both zero-shot and CoT prompting require only about 9 seconds.

\begin{figure}[h]
    \centering
    \begin{subfigure}[]{1.0\linewidth}
        \centering
        \includegraphics[width=1.0\linewidth]{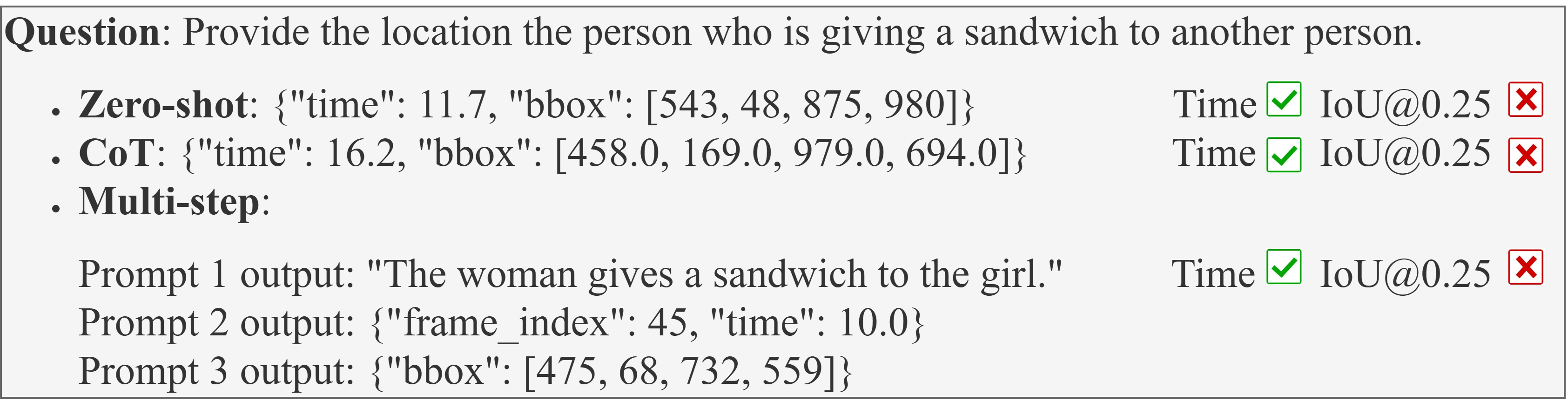}
        \subcaption{}
        \label{}
    \end{subfigure}
    \hfill
    \begin{subfigure}[]{1.0\linewidth}
        \centering
        \includegraphics[width=1.0\linewidth]{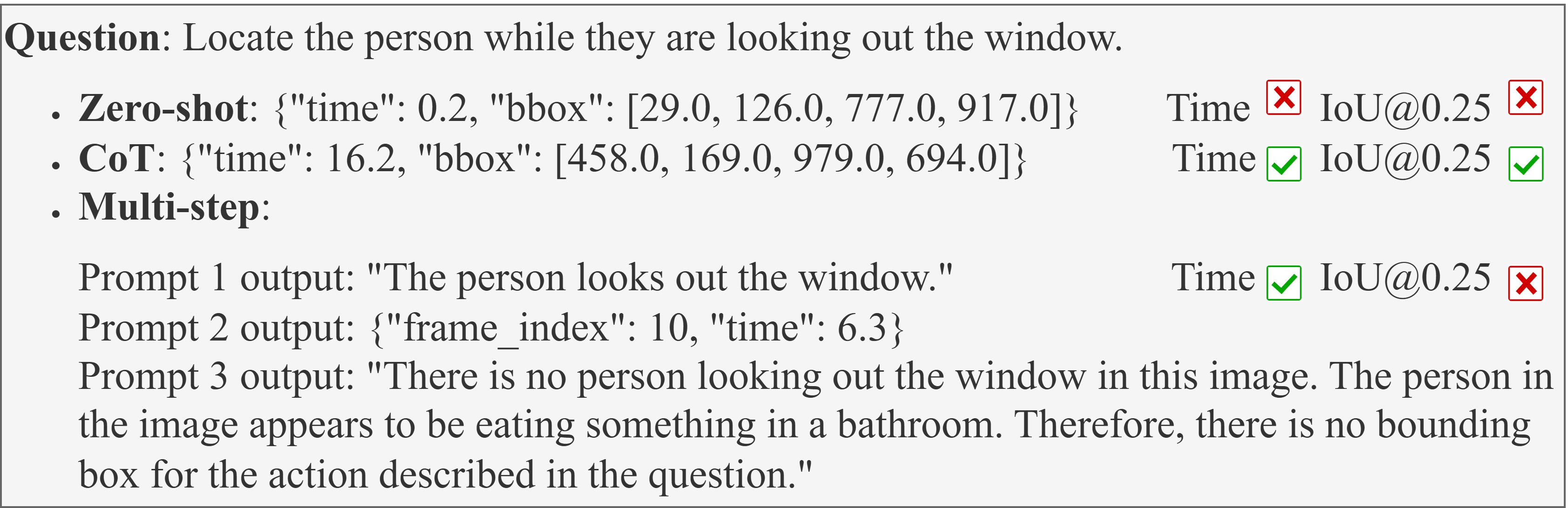}
        \subcaption{}
        \label{}
    \end{subfigure}
    % \hfill
    % \begin{subfigure}[]{1.0\linewidth}
    %     \centering
    %     \includegraphics[width=0.7\linewidth]{images/raw_outputs/prompt_raw_3.png}
    %     \subcaption{}
    %     \label{}
    % \end{subfigure}
    \caption{Raw outputs Qwen under different prompting strategies. }
    \label{fig:raw_output_prompts}
\end{figure}

We also present representative raw model outputs under the three prompting configurations in Fig.~\ref{fig:raw_output_prompts}. These examples illustrate how the model organizes its intermediate reasoning and final predictions under each setting. Despite slight differences in some cases, the final predictions remain largely consistent across the three prompting strategies.

% \begin{figure}[h]
%   \centering
%   \includegraphics[width=\linewidth]{images/prompts}
%   \caption{Raw model outputs for Zero-shot, CoT and multi-step reasoning in QwenVL 2.5 model}
%   \label{fig:raw_output_prompts}
% \end{figure}

\begin{table*}[h]

\centering
\resizebox{\textwidth}{!}{

\begin{tabular}{|l|
                c |c| >{\cellcolor{gray!15}}c|
                c |c| >{\cellcolor{gray!15}}c|
                c |c |c| >{\cellcolor{gray!15}}c|
                c |c |c |c| >{\cellcolor{gray!15}}c|
                c| c| c|>{\cellcolor{gray!15}}c |
                c|}

\hline
\multirow{2}{*}{Model} & 
\multicolumn{3}{c|}{PGR} &
\multicolumn{3}{c|}{OGR} &
\multicolumn{4}{c|}{POCoGR} &
\multicolumn{5}{c|}{HOCoGR} &
\multicolumn{4}{c|}{TGR} &
\multirow{2}{*}{Inference Time} \\

\cline{2-20}

& $T_{Acc}$ & $PIoU$ & $Acc^{SGR}$
& $T_{Acc}$ & $OIoU$ & $Acc^{SGR}$
& $T_{Acc}$ & $PIoU$ & $OIoU$ & $Acc^{SGR}$
& $T_{Acc}$ & $LHIoU$ & $RHIoU$ & $OIoU$ & $Acc^{SGR}$
& $A_{Acc}$ & $MAD$ & $T@2$ & $Acc^{TGR}$ &\\

\hline

Zero-shot 
& 17.1 & 33.6 & 5.5  
& 20.8 & 7.8 & 1.4 
& 23.8 & 43.0 & 10.2 & 2.2 
& 2.7 & 2.4 & 0.3 & 0.9 & 0.0 
& 50.5 & 6.4 & 26.9 & 16.9 & 9.1\\

CoT 
& 19.8 & 29.4 & 6.3 
& 24.0 & 5.6 & 2.2 
& 26.8 & 42.4 & 9.1 & 2.0 
& 2.5 & 3.5 & 2.0 & 1.5 & 0.0 
& 55.8 & 6.6 & 19.4 & 10.9 & 9.2\\

Multi-step 
& 24.7 & 33.9 & 7.8 & 21.6 & 11.6 & 2.2 & 23.9 & 35.5 & 10.8 & 1.6
& 6.2 & 0.0 & 0.3 & 0.3 & 0.0 & 61.2 & 13.3 & 16.2 & 11.2 & 20.8\\

\hline
\end{tabular}
}
\caption{Zero-shot, CoT, and multi-step inference in Qwen2.5 VL 7B.}
\label{tab:ablation_step_by_step}
\end{table*}

% ###############################################################################################################

\section{Ablation Studies of GRAM on Input Resolution and Number of Input Frames}
\label{sup:ablations}

\subsection{Effect of Input Resolution}
\label{sup:ablation_resolution}

Table~\ref{tab:ablation_video_resolution} analyzes the effect of increasing the input spatial resolution (long-side resolution) across the five reasoning scenarios while keeping the number of input video frames fixed at 24. Overall, higher resolution improves overall performance in spatial grounded reasoning tasks, 
% indicating that finer spatial details are particularly important for accurate object localization. The improvements are especially evident in the POCoGR scenario, suggesting that resolving interactions between multiple entities benefits from higher spatial resolution.
indicating that finer spatial details are particularly important for accurate object localization and resolving interactions between multiple entities.
% benefits from higher spatial resolution.
In contrast, hand–object co-grounded reasoning seems highly challenging even at higher resolutions.
% , highlighting the difficulty of accurately detecting small and highly articulated entities such as hands. 
Temporal grounded reasoning also demonstrates gains, suggesting that clearer spatial cues can also assist relative and absolute temporal reasoning and localization. Due to computational constraints, the maximum resolution we evaluate is 512, which we adopt as the default resolution in GRAM.

\begin{table*}[h]

\centering
\resizebox{\textwidth}{!}{

\begin{tabular}{|l|
                c |c| >{\cellcolor{gray!15}}c|
                c |c| >{\cellcolor{gray!15}}c|
                c |c |c| >{\cellcolor{gray!15}}c|
                c |c |c |c| >{\cellcolor{gray!15}}c|
                c| c| c|>{\cellcolor{gray!15}}c |}

\hline
\multirow{2}{*}{Resolution} & 
\multicolumn{3}{c|}{PGR} &
\multicolumn{3}{c|}{OGR} &
\multicolumn{4}{c|}{POCoGR} &
\multicolumn{5}{c|}{HOCoGR} &
\multicolumn{4}{c|}{TGR} \\

\cline{2-20}

& $T_{Acc}$ & $PIoU$ & $Acc^{SGR}$
& $T_{Acc}$ & $OIoU$ & $Acc^{SGR}$
& $T_{Acc}$ & $PIoU$ & $OIoU$ & $Acc^{SGR}$
& $T_{Acc}$ & $LHIoU$ & $RHIoU$ & $OIoU$ & $Acc^{SGR}$
& $MAD$ & $A_{Acc}$ & $T@2$ & $Acc^{TGR}$ \\

\hline

128 & 25.7 & 21.3 & 8.0 & 48.0 & 3.9 & 7.9 & 43.8 & 19.4 & 11.4 & 3.7 & 4.2 & 0.0 & 0.0 & 0.0 & 0.0 & 43.4 & 22.0 & 32.9 & 17.8 \\
224 & 28.4 & 51.3 & 14.9 & 50.8 & 23.8 & 13.4 & 45.9 & 42.7 & 17.7 & 5.7 & 7.0 & 0.3 & 0.0 & 0.0 & 0.0 & 46.0 & 21.9 & 32.7 & 19.0 \\
336 & 29.6 & 75.5 & \underline{22.7} & 52.2	& 33.8	& \textbf{22.0} & 43.5 & 70.3 & 29.7 & \underline{11.4} & 6.5 & 0.3 & 0.0 & 0.0 & 0.0 & 46.9 & 20.9 & 34.7 & \underline{22.3} \\
512 (GRAM) & 33.4 & 80.1 & \textbf{26.6} & 50.2 & 37.0 & \underline{21.2} & 39.3 & 80.7 & 73.1 & \textbf{28.0} & 3.5 & 5.8 & 0.0 & 0.0 & 0.0 & 48.7 & 22.1 & 36.1 & \textbf{23.7} \\

\hline
\end{tabular}
}
\caption{Ablation study on effect of input resolution.}
\label{tab:ablation_video_resolution}
\end{table*}

\subsection{Effect of Number of Input Frames}
\label{sup:ablation_num_frames}

% \blue{
Table~\ref{tab:ablation_video_frames} analyzes the effect of increasing the number of input video frames, while keeping the input resolution fixed at 512, which determines the temporal context available to the model. 
Overall, increasing the number of frames consistently improves performance across most reasoning scenarios, indicating that richer temporal context helps the model better capture action dynamics and associate entities with the correct temporal segments.
Notably, TGR clearly benefits from additional frames, as a broader temporal window enables the model to more accurately determine both the relative and absolute temporal extent of actions. Improvements are also observed in PGR, OGR, and POCoGR, suggesting that longer temporal context helps disambiguate which entities participate in an action and enhances the joint reasoning and localization of interacting entities.
Considering the overall improvements and computational trade-offs, we adopt 24 frames as the default temporal context in GRAM.
% }

\begin{table*}[h]

\centering
\resizebox{\textwidth}{!}{

\begin{tabular}{|l|
                c |c| >{\cellcolor{gray!15}}c|
                c |c| >{\cellcolor{gray!15}}c|
                c |c |c| >{\cellcolor{gray!15}}c|
                c |c |c |c| >{\cellcolor{gray!15}}c|
                c| c| c|>{\cellcolor{gray!15}}c |}

\hline
\multirow{2}{*}{\# video frames} & 
\multicolumn{3}{c|}{PGR} &
\multicolumn{3}{c|}{OGR} &
\multicolumn{4}{c|}{POCoGR} &
\multicolumn{5}{c|}{HOCoGR} &
\multicolumn{4}{c|}{TGR} \\

\cline{2-20}

& $T_{Acc}$ & $PIoU$ & $Acc^{SGR}$
& $T_{Acc}$ & $OIoU$ & $Acc^{SGR}$
& $T_{Acc}$ & $PIoU$ & $OIoU$ & $Acc^{SGR}$
& $T_{Acc}$ & $LHIoU$ & $RHIoU$ & $OIoU$ & $Acc^{SGR}$
& $MAD$ & $A_{Acc}$ & $T@2$ & $Acc^{TGR}$ \\

\hline

8         & 33.1 & 75.0 & 24.0 & 48.2 & 36.2 & \textbf{21.2} & 41.5 & 0.0 & 0.0 & 0.0    & 2.7 & 0.3 & 0.0 & 0.3 & 0.0 & 44.5 & 21.7 & 26.3 & 15.3 \\
16        & 31.6 & 75.9 & \underline{24.3} & 47.8 & 37.8 & \underline{21.1} & 32.4 & 0.0 & 0.0 & 0.0    & 3.5 & 0.5 & 0.0 & 0.0 & 0.0 & 47.3 & 20.9 & 28.6 & \underline{15.8} \\
24 (GRAM) & 33.4 & 80.1 & \textbf{26.6} & 50.2 & 37.0 & \textbf{21.2} & 39.3 & 80.7 & 73.1 & \textbf{28.0} & 3.5 & 5.8 & 0.0 & 0.0 & 0.0 & 48.7 & 22.1 & 36.1 & \textbf{23.7} \\

\hline

\end{tabular}
}
\caption{Ablation study on temporal context.}
\label{tab:ablation_video_frames}
\end{table*}

% ###############################################################################################################

\section{Visual Ablation Study of GRAM for Spatio-Temporal Grounded Reasoning}
\label{sup:gram_reasoning}

\begin{figure*}[ht!]
  \centering
  \includegraphics[width=0.9\textwidth]{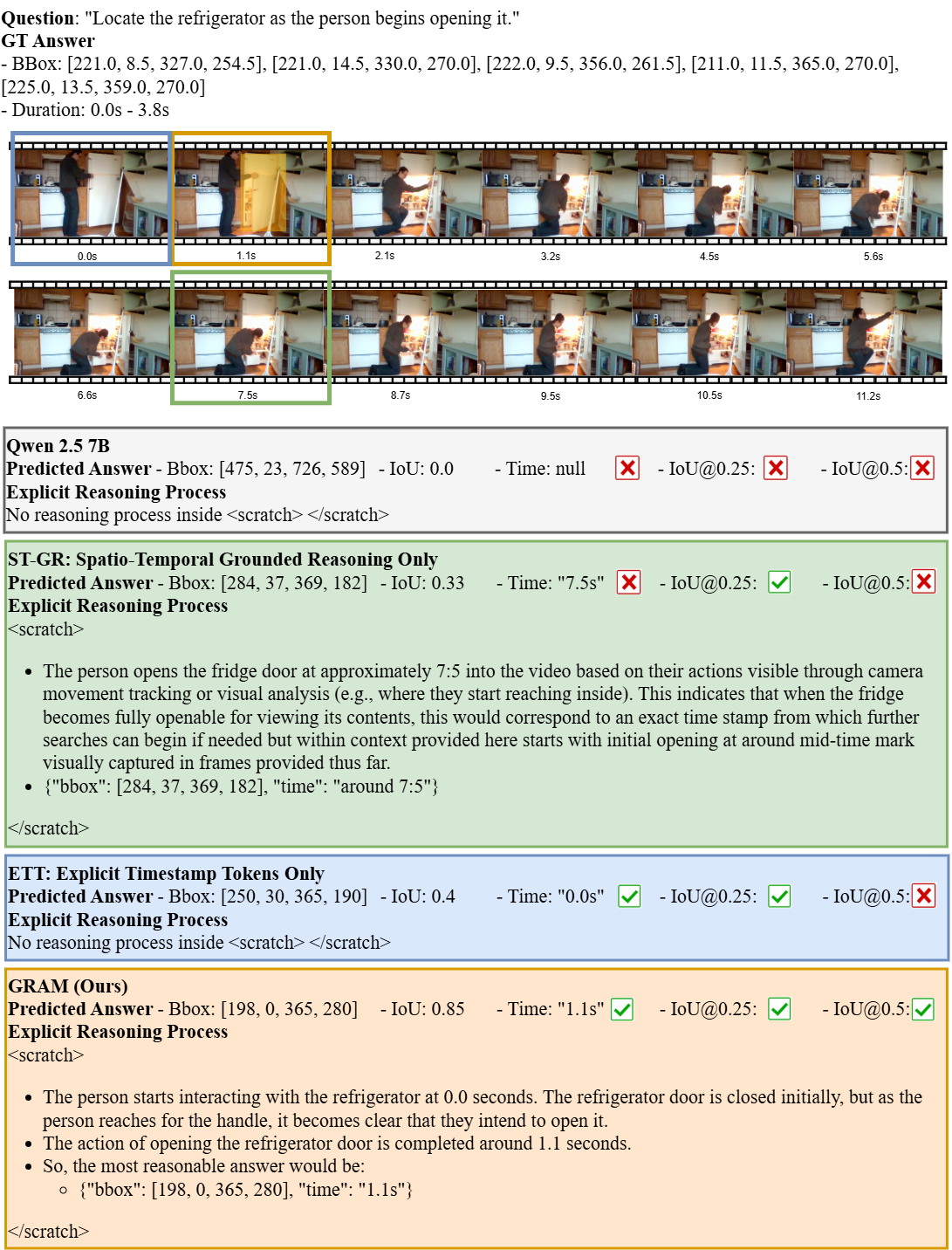}
  \caption{GRAM's spatio-temporal grounded reasoning process under Object Grounded Reasoning category.}
  \label{fig:gram_reasoning}
\end{figure*}

In Table \ref{tab:ablation} of the main paper, evaluate the contributions of each component in the GRAM framework.
Specifically, we compare four variants: (A) base Qwen 2.5 VL 7B model, (B) spatio-Temporal Grounded Reasoning (ST-GR),
where each reasoning step produced by the model is explicitly grounded by relevant spatio-temporal evidence in the video,
(C) Explicit Timestamp Tokens (ETT) where we interleave absolute time tokens with frame embeddings in the input sequence, and (D) the full GRAM model, which combines both (B) and (C).
Figure~\ref{fig:gram_reasoning} provides a qualitative visualization of this ablation. 
As shown in the example, 
Qwen struggles in this example: it fails to localize the \textit{refrigerator} and does not produce a valid action timestamp. When ST-GR is added, the model becomes able to localize the \textit{refrigerator} and exhibits partial spatial grounding behavior in its reasoning steps. However, it still fails to bind this spatial grounding to the correct action duration, predicting a starting time of 7.5 seconds, at which point the person has already opened the refrigerator and begun taking food out. Moreover, although ST-GR improves spatial grounding, the model still fails under the stricter IoU threshold of 0.5. 
Adding ETT alone leads to complementary improvements: the model correctly identifies the absolute starting time of the action and also localizes the refrigerator, though only at a lower IoU threshold of 0.25.
When both mechanisms, ST-GR and ETT are combined, GRAM produces the most coherent and accurate reasoning trajectory. It successfully localizes the refrigerator, aligns the spatial grounding with the appropriate action duration, maintains robust detection even at higher IoU thresholds, achieving 0.85 (as shown in the 1.1s frame outlined in yellow, with the transparent yellow bounding box).
This example qualitatively demonstrates that ST-GR strengthens the model’s ability to focus on the most relevant spatial evidence, while ETT ensures that the grounded visual cues are temporally aligned with the correct absolute timestamps. Their combination enables GRAM to achieve reliable spatio-temporal grounded reasoning.

% \begin{figure*}[t]
%   \centering
%   \includegraphics[width=0.78\textwidth]{images/attention_visualization/att_visualization_person-object_444.png}
%   \caption{GRAM's spatio-temporal grounded reasoning process under Person-Object Co-Grounded Reasoning category.}
%   \label{fig:gram_reasoning2}
% \end{figure*}

% #################################################################################
% \section{Effect of larger models}
% \label{sup:ablation_num_frames}

% ###############################################################################################################

% ###############################################################################################################

% \section{Qualitative Results}
% \label{sup:qualitative_results}

% % \red{include visualization for videos with more than one person}
% % \red{include reasoning traces also}

% In Section~\ref{supp:qualitative_comparision}, we present qualitative results for all evaluated models, including GRAM, across the five test scenarios.
% In Section~\ref{supp:gram_reasoning}, we visualize the spatio-temporal grounded reasoning behaviour of GRAM.
% % \red{In Section~\ref{sup:GRAM_fail}, we provide several failure cases of GRAM on the Know-Show test scenarios.}

\section{Qualitative Comparison of Models on Know-Show Test Scenarios}
\label{sup:qualitative_results}

\begin{figure*}[t]
  \centering
  \includegraphics[width=0.8\textwidth]{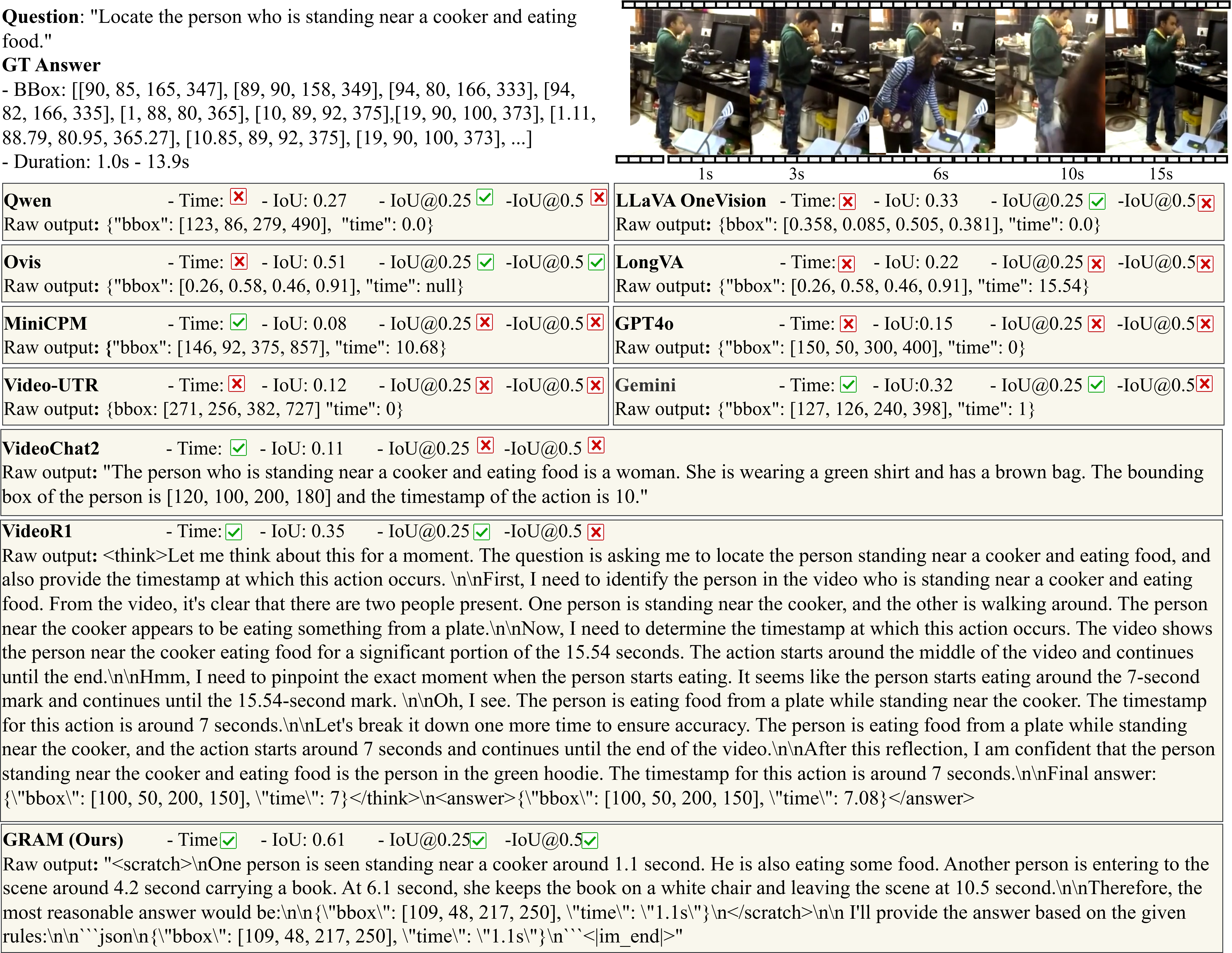}
  \caption{Qualitative results for Action Conditioned Person Grounded Reasoning.}
  \label{fig:person}
\end{figure*}

% \blue{
In this section, we present qualitative results for all evaluated models, including GRAM, across the five test scenarios.
In Fig.~\ref{fig:person}, we present qualitative results for the Action-Conditioned Person Grounded Reasoning scenario. Along with the model predictions, we also include the raw outputs generated by each Video-LM. The video contains two individuals: one standing near the cooking area eating food, and another entering the scene, placing a book on a chair, and leaving.
Qwen, Ovis, and LLaVA-OneVision correctly identify the relevant person but fail to associate the prediction with the correct temporal action duration. In contrast, MiniCPM and VideoChat2 correctly predicts the time but fails to localize the person spatially. GRAM, VideoR1, and Gemini successfully predict both the time and the spatial region at IoU@0.25. However, under stricter IoU thresholds, only GRAM maintains consistent performance, while VideoR1 and Gemini fail to preserve accurate localization.
Another notable observation concerns the reasoning traces produced by different models. GRAM, VideoR1 and VideoChat2 explicitly provide intermediate reasoning steps before producing the final prediction. In contrast, the other models directly output the final result in a JSON structure containing exactly two keys "bbox" and "time", as specified in the prompt. This suggests that these models may perform internal reasoning that is not exposed to the user, returning only the final prediction.
Comparing the reasoning traces, it seems that GRAM’s explanations are concise and well-grounded in the visual evidence, whereas VideoR1 produces overly verbose outputs with noticeable repetition. Nevertheless, the reasoning trajectory of VideoR1 remains logically correct. 
Gemini, on the other hand, does not provide intermediate reasoning but still produces the correct prediction.
% }

\begin{figure*}[t]
  \centering
  \includegraphics[width=0.8\textwidth]{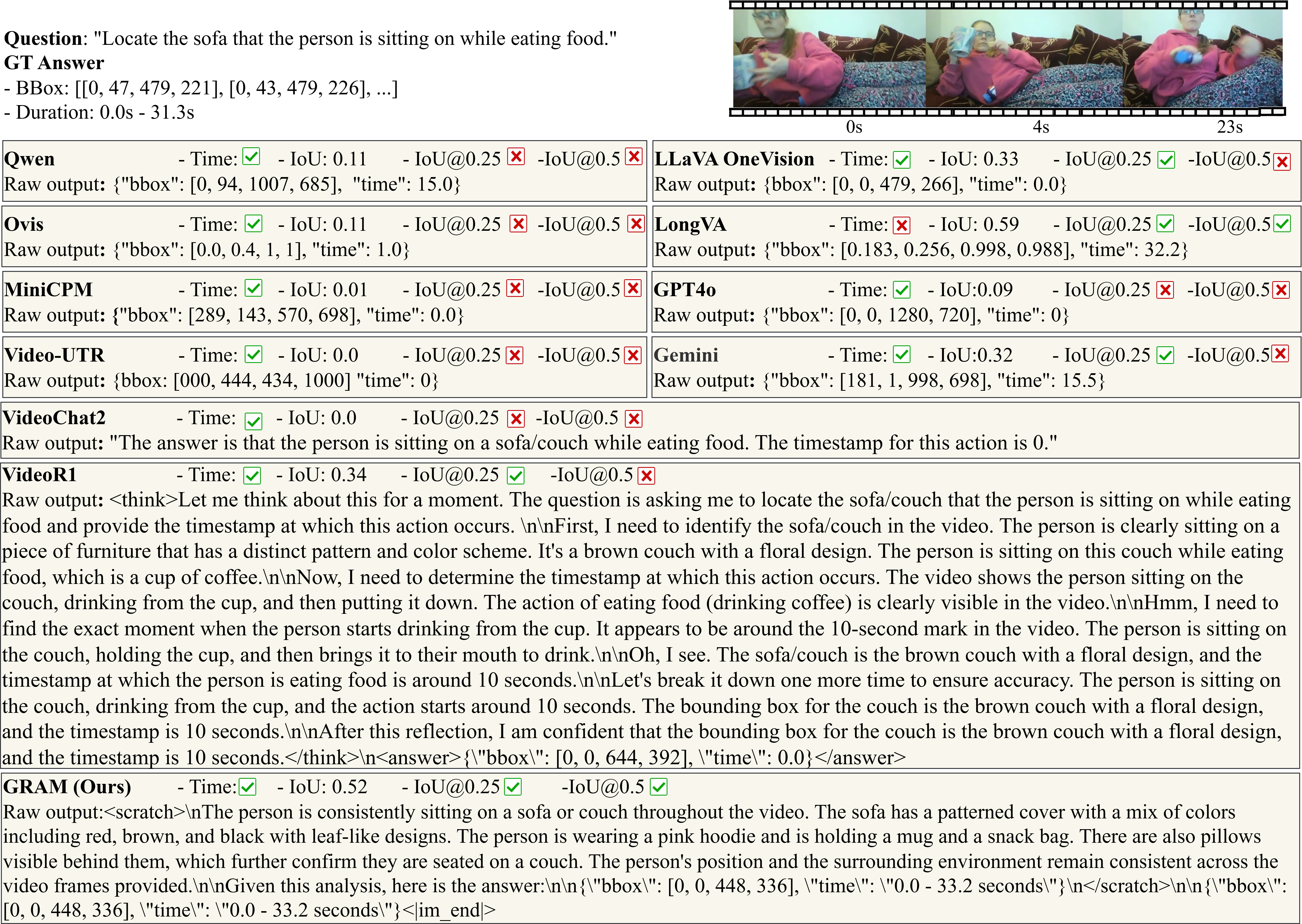}
  \caption{Qualitative results for Action Conditioned Object Grounded Reasoning.}
  \label{fig:obj}
\end{figure*}

% \blue{
In Fig.~\ref{fig:obj}, we present qualitative results for the Action-Conditioned Object Grounded Reasoning scenario. 
While most models, correctly identify the temporal interval associated with the queried action, object grounding remains challenging. 
Only VideoR1, LLaVA-OneVision, Gemini and GRAM successfully identify both the relevant object and the correct temporal grounding, even though the object \textit{sofa} remains stationary and clearly visible throughout the video.
% Except for GRAM, VideoR1 and LLaVA-OneVision, all other models miss this object entirely. 
Furthermore, when evaluated under a stricter localization criterion, only GRAM maintains correct grounding at IoU@0.5. 
Regarding reasoning traces, both GRAM and VideoR1 provide intermediate explanations that are logically consistent and grounded in visual evidence, whereas the reasoning output produced by VideoChat2 primarily focuses on the temporal aspect.
% }

\begin{figure*}[t]
  \centering
  \includegraphics[width=0.81\textwidth]{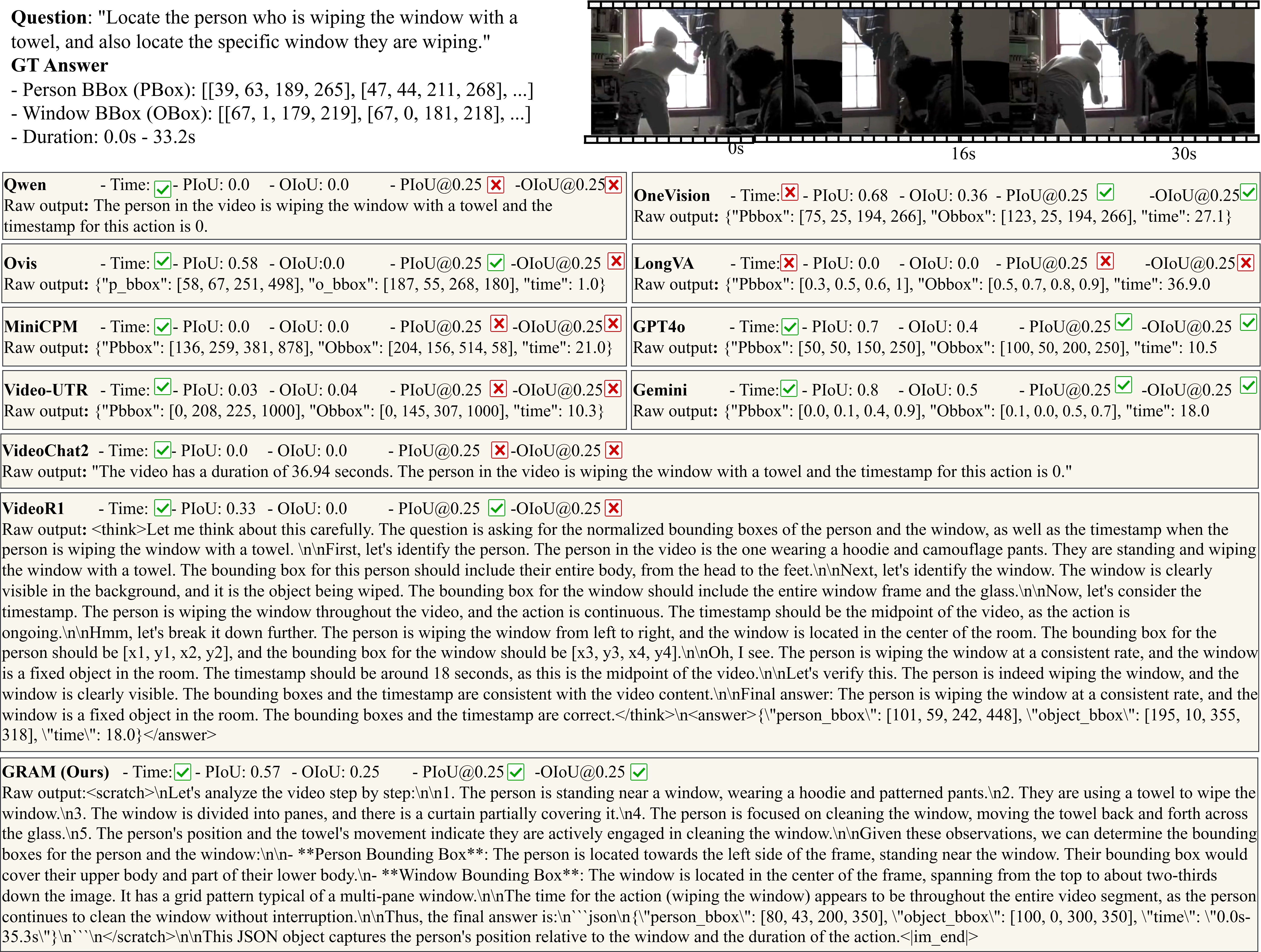}
  \caption{Qualitative results for Action Conditioned Person-Object Co-Grounded Reasoning.}
  \label{fig:person-obj}
\end{figure*}

% \blue{
In Fig.~\ref{fig:person-obj}, we present qualitative results for the Action-Conditioned Person–Object Co-Grounded Reasoning scenario. Most models including Qwen, Ovis, MiniCPM, Video-UTR, Video-R1, GPT-4o, Gemini, and GRAM correctly identify the temporal interval associated with the queried action. However, accurate spatial grounding and actor disambiguation seems more challenging. Only GPT-4o, Gemini, and GRAM successfully localize both the relevant person and the associated object in space and time, while also distinguishing the correct person performing the queried action among the two individuals present in the scene. Although Ovis and Video-R1 correctly identify the person and the temporal segment, they fail to detect the associated object.
% }

\begin{figure*}[!t]
  \centering
  \includegraphics[width=0.81\textwidth]{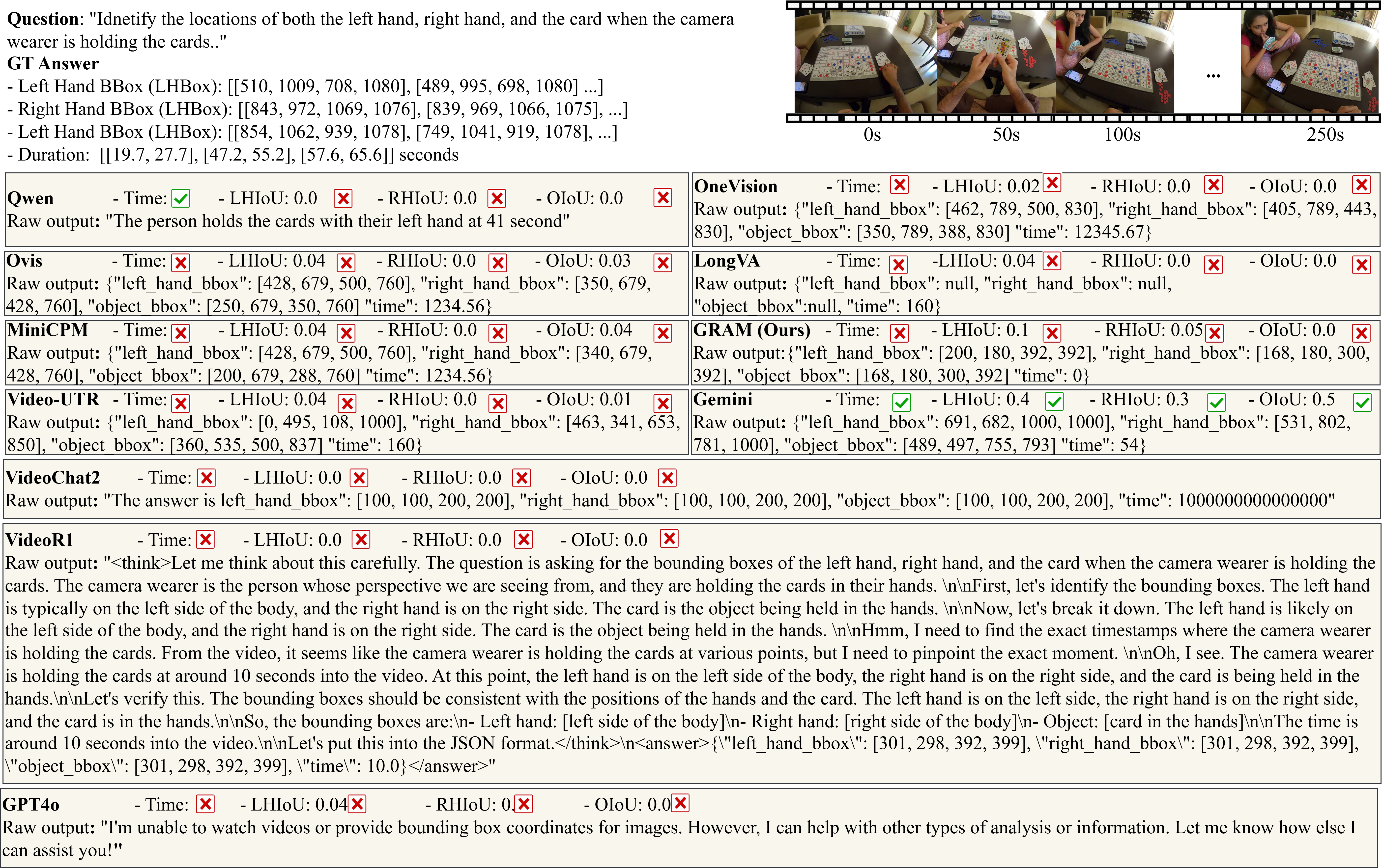}
  \caption{Qualitative results: Action Conditioned Hand-Object Co-Grounded Reasoning.}
  \label{fig:hand-obj}
\end{figure*}

% \blue{
In Fig.~\ref{fig:hand-obj}, we present qualitative results for the Action-Conditioned Hand–Object Co-Grounded Reasoning scenario. Among all evaluated spatial gounded reasoning settings, this represents the most challenging case, as the model must simultaneously identify the left hand, right hand, and the manipulated object, while correctly associating them with the relevant temporal segment of the action. As shown in the figure, only Gemini successfully localizes both the left hand and the object while aligning them with the correct time span.
% }

\begin{figure*}[!t]
  \centering
  \includegraphics[width=0.9\textwidth]{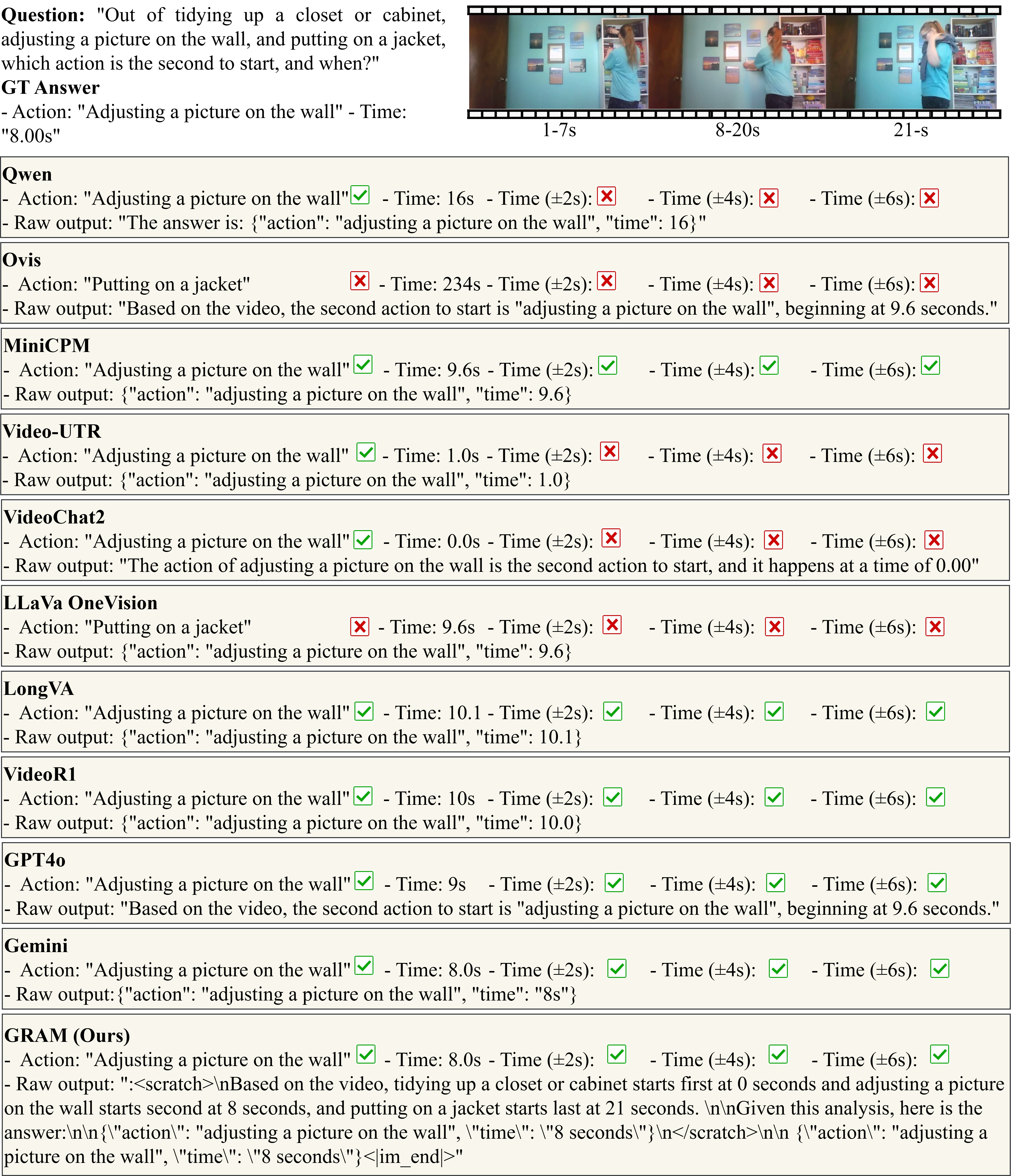}
  \caption{Qualitative results for Temporal Grounded Reasoning.}
  \label{fig:temporal}
\end{figure*}

% \blue{
In Fig.~\ref{fig:temporal}, we demonstrate the temporal grounded reasoning capabilities of the models. In this scenario, most models, including Qwen, MiniCPM, LongVA, VideoR1, GPT-4o, Gemini, and GRAM, correctly identify both the relative temporal order of the actions and their absolute starting timestamps within the video. Although Video-UTR and VideoChat2 successfully capture the relative ordering of events, they fail to localize the starting time of the action even with a grace window of 6 seconds. Both forms of temporal grounding are essential, as robust video understanding requires models to reason not only about when events occur relative to one another but also their precise positions along the video timeline.

\end{document}